\documentclass[a4paper,10pt,reqno]{amsart}
\usepackage{times}

\usepackage{amsmath,amsfonts,bm}

\def\eqref#1{equation~\ref{#1}}

\def\1{\bm{1}}

\def\rvx{{\mathbf{x}}}

\def\rvz{{\mathbf{z}}}

\DeclareMathAlphabet{\mathsfit}{\encodingdefault}{\sfdefault}{m}{sl}
\SetMathAlphabet{\mathsfit}{bold}{\encodingdefault}{\sfdefault}{bx}{n}

\def\gC{{\mathcal{C}}}

\def\gE{{\mathcal{E}}}

\def\gK{{\mathcal{K}}}

\def\gN{{\mathcal{N}}}

\def\gP{{\mathcal{P}}}
\def\gQ{{\mathcal{Q}}}

\def\gT{{\mathcal{T}}}

\def\gV{{\mathcal{V}}}
\def\gW{{\mathcal{W}}}

\def\gZ{{\mathcal{Z}}}

\def\sP{{\mathbb{P}}}

\newcommand{\E}{\mathbb{E}}

\newcommand{\R}{\mathbb{R}}

\DeclareMathOperator*{\argmin}{arg\,min}

\DeclareMathOperator{\Tr}{Tr}

\usepackage{lipsum}
\usepackage{amsfonts}
\usepackage{amsthm}
\usepackage{graphicx}
\usepackage{epstopdf}
\usepackage{soul}
\usepackage{xcolor}
\usepackage{algorithm}
\usepackage{algpseudocode}
\usepackage{subcaption}
\usepackage{mathtools}
\usepackage{booktabs}
\usepackage{cases} 

\usepackage{hyperref}
\usepackage{cleveref}
\usepackage{url}

\usepackage{natbib}
\usepackage{fancyhdr}
 \usepackage[foot]{amsaddr}

\newtheorem{theorem}{Theorem}
\newtheorem{definition}{Definition}
\newtheorem{proposition}{Proposition}
\newtheorem{corollary}{Corollary}
\newtheorem{remark}{Remark}
\DeclareMathOperator{\wprox}{WProx}
\DeclareMathOperator{\diag}{diag}

\DeclareMathOperator{\TV}{TV}

\newcommand{\cc}{}
\newcommand{\ttS}{\tilde\Sigma}

\def\addcontentsline#1#2#3{}

\title{Noise-Free Sampling Algorithms via Regularized Wasserstein Proximals}

\author[Tan]{Hong Ye Tan} 
\email{hyt35@cam.ac.uk}
\address[A1]{Department of Applied Mathematics and Theoretical Physics, University of Cambridge.}
\author[Osher]{Stanley Osher} 
\email{sjo@math.ucla.edu}
\address[A2]{Department of Mathematics, University of California, Los Angeles, 90095}
\author[Li]{Wuchen Li}
\email{wuchen@mailbox.sc.edu}
\address[A3]{Department of Mathematics, University of South Carolina, Columbia, SC 29208.}

\begin{document}

\maketitle
\begin{abstract}
We consider the problem of sampling from a distribution governed by a potential function. This work proposes an explicit score based MCMC method that is deterministic, resulting in a deterministic evolution for particles rather than a stochastic differential equation evolution. The score term is given in closed form by a regularized Wasserstein proximal, using a kernel convolution that is approximated by sampling. We demonstrate fast convergence on various problems and show improved dimensional dependence of mixing time bounds for the case of Gaussian distributions compared to the unadjusted Langevin algorithm (ULA) and the Metropolis-adjusted Langevin algorithm (MALA). We additionally derive closed form expressions for the distributions at each iterate for quadratic potential functions, characterizing the variance reduction. Empirical results demonstrate that the particles behave in an organized manner, lying on level set contours of the potential. Moreover, the posterior mean estimator of the proposed method is shown to be closer to the maximum a-posteriori estimator compared to ULA and MALA in the context of Bayesian logistic regression. Additional examples demonstrate competitive performance for Bayesian neural network training.
\end{abstract}

\section{Introduction}
Sampling from an unknown distribution is a fundamental task in data science. Notable applications include maximum likelihood estimation and uncertainty quantification \citep{laumont2022bayesian}, Bayesian neural networks training \citep{mackay1995bayesian}, global optimization \citep{dai2021global}, and generative modelling \citep{batzolis2021conditional,hyvarinen2005estimation,song2020score}. In general, the problem can be formulated as sampling from a Gibbs distribution, with a density of the form
\begin{equation*}
    \rho(x) \sim \exp(-V(x)/\beta),
\end{equation*}
where $\beta$ is a regularization parameter, and $V$ is a known bounded $\gC^1$ potential function, satisfying appropriate growth conditions such that $\rho$ is a well defined density function. One popular way to do this is using Markov chain Monte Carlo (MCMC) algorithms \citep{andrieu2003introduction,brooks2011handbook}. MCMC algorithms work by first constructing a Markov chain whose stationary distribution is equal or close to the target distribution. By using ergodic theory, the Markov chains can be shown to converge in distribution from a tractable initial distribution to the intractable stationary distribution. Hence, to sample from the target distribution, one needs only evaluate the Markov chain for a suitably large number of iterations.

There are three main paradigms for MCMC: zeroth order methods, first order methods, and score based methods. Some examples of zeroth order methods include the Metropolized random walk and hit-and-run algorithms, which do not use the gradient of the potential $\nabla V$ \citep{mengersen1996rates,belisle1993hit}. First order methods utilize the gradient of our potential $V$ as well as randomness to converge in distribution to the target distribution. Two of the most popular first order methods are the unadjusted Langevin algorithm (ULA) and the Metropolis-adjusted Langevin algorithm (MALA) \citep{parisi1981correlation,durmus2019high,rossky1978brownian,brooks2011handbook}. These two algorithms were subsequently extended using modifications including acceleration \citep{wang2022accelerated}, proximal steps \citep{pereyra2016proximal}, Riemannian metrics \citep{patterson2013stochastic}, Hamiltonians \citep{betancourt2017conceptual}, and projections \citep{wang2022accelerated}. ULA and MALA consider discretizing an SDE that corresponds to the Fokker-Planck equation. Many common zeroth and first order sampling methods, including ULA and MALA, rely on randomness that is independent of the samples to guarantee ergodicity of the Markov chains, typically modelled using white Gaussian noise. This randomness generates sufficient diffusion, which is then used show convergence \citep{mattingly2002ergodicity,meyn1994computable}. 

While diffusion can be achieved using random noises, we instead consider the third paradigm of achieving diffusion using the score of the density $\nabla \log \rho(x)$. Score based methods reformulate the Fokker-Planck equation into an ODE instead of an SDE, with the ODE depending on the gradient of the log-likelihood (the score) of the density \citep{maoutsa2020interacting,song2020score,del2013mean}. Some recent applications of score based diffusion include conditional generative modelling, utilizing the backwards Kolmorogov equation to diffuse from noise to natural images \citep{song2020score,batzolis2021conditional}. However, the score is not available, as it depends on the target density. Various methods have been proposed to approximate the score, including kernel density estimation \citep{carrillo2019blob,terrell1992variable,kim2012robust,wand1994kernel}, adaptive kernel methods \citep{van2003adaptive,botev2010kernel}, and neural ODEs \citep{bond2021deep, chen2018neural,nijkamp2022mcmc}. These approaches are generally non-parametric, without making a-priori assumptions on the target distribution. However, such approximations face common problems such as choice of kernel, mode collapse and sensitivity to hyper-parameters \citep{srivastava2017veegan,li2023reducing,gramacki2018nonparametric}. We propose an alternative formulation of score approximation using the approximate Wasserstein proximal of the empirical measure, with a principled method of choosing hyper-parameters, that produces samples from a modified density that is close to the target density.

Utilizing Liouville's equation, we consider a score ODE to be solved in the particle space, whose density evolves according to the Fokker-Planck equation. The Jordan-Kinderlehrer-Otto (JKO) scheme considers a discretization of the Fokker-Planck ODE using proximal mappings in the Wasserstein space \citep{jordan1998variational}. The target ODE is of the form
\begin{equation*}
    \frac{dX}{dt} = -\nabla V(X) - \beta\nabla \log \rho(t,X),
\end{equation*}
where $\rho(t)$ is the density of $X_t$ at time $t$. The JKO scheme discretizes the ODE using Wasserstein proximal operators of the form 
\begin{equation*}
    \rho_{k+1} = \argmin_{\rho \in \gP_2} \int_{\R^d} (\beta \rho \log \rho + V \rho) dx + \frac{1}{2h} \gW(\rho_k,\rho)^2,
\end{equation*}
where $k$ is the iteration of the update, $h>0$ is the step-size, $\gP_2$ is the space of probability densities over $\R^d$ with bounded second moments, and $\gW$ is the Wasserstein-2 distance between probability measures. The JKO scheme is the proximal iteration for the free energy functional with the Wasserstein-2 metric. However, the proximal map of the density is generally intractable and requires solving an equivalently difficult problem to our sampling problem. A recent work has considered using a regularized proximal term, formulated in terms of a set of coupled forward- and backward-heat equations \citep{li2023kernel}. The score of the regularized Wasserstein proximal term has a \emph{closed-form} solution based on convolutions with heat kernels. Motivated by this, we propose to utilize the closed-form solution for deterministic sampling. 

In this work, we propose a deterministic sampling method based on the score flow. We then demonstrate stable convergence as well as convergence in the case of Gaussian densities, where we demonstrate a better dimension dependence bound due to the closed form solution in this case. Our proposed method is then compared with the unadjusted Langevin algorithm (ULA) as well as the Metropolis-adjusted Langevin algorithm (MALA), which are both stochastic methods. In the rest of this section, we introduce the Fokker-Planck equation, as well as the associated SDE and score ODE. 

\subsection{Definitions}
We begin with some preliminary definitions, including the Wasserstein distance metric between probability measures, as well as the Fokker-Planck equation.

\begin{definition}
    For two probability density functions $\mu, \eta$ on $\R^d$ with finite second moment, the \emph{Wasserstein-2} distance between $\mu$ and $\eta$ is
    \begin{equation*}
        \gW(\mu, \eta) \coloneqq \left(\inf_{\gamma \in \Gamma(\mu, \eta)} \iint_{\R^d\times \R^d} \|x-y\|^2 \gamma(x,y)\, dx\, dy \right)^{1/2},
    \end{equation*}
    where the norm is the Euclidean norm, and the infimum is taken over all couplings $\gamma \in \Gamma(\mu, \eta)$ between $\mu$ and $\eta$, i.e. $\gamma$ is a joint probability measure on $\R^d \times \R^d$ with
    \[\int_{\R^d} \gamma(x,y)\, dy = \mu(x),\quad \int_{\R^d} \gamma(x,y)\, dx = \eta(y).\]
    
    Let $\rho_0$ be a probability density function with finite second moment, and $V \in \mathcal{C}^1(\R^d)$ be a bounded potential function. For a scalar $T>0$, the \emph{Wasserstein proximal} of $\rho_0$ is defined as 
    \begin{equation}
        \rho_T = \wprox_{TV}(\rho_0) \coloneqq \argmin_{q\in \mathcal{P}_2(\R^d)} \int_{\R^d} V(x)q(x) d x + \frac{\gW(\rho_0, q)^2}{2T},
    \end{equation}
    where $\gW(\rho_0, q)$ is the Wasserstein-2 distance between $\rho_0$ and $q$, and $\gP_2$ is the set of probability density functions $q$ with finite second moment.
\end{definition}

The Wasserstein proximal does not admit an easily computable solution, and thus we consider an approximation to the Wasserstein proximal. \citet{li2023kernel} consider an optimal control formulation based on the Benamou-Brenier formula \citep{benamou2000computational}. This reformulates the variational problem into a coupled ODE system. The regularized Wasserstein proximal operator is thus defined by the solution of the regularized PDEs
\begin{subequations}\label{eqs:regPDE}
    \begin{numcases}{}
      \partial_t \rho(t,x) + \nabla_x \cdot\left(\rho(t,x) \nabla_x \Phi(t,x)\right) = \beta \Delta_x \rho(t,x), \label{eq:regPDE_a} \\
      \partial_t \Phi(t,x) + \frac{1}{2} \|\nabla_x \Phi(t,x)\|^2 = -\beta \Delta_x \Phi(t,x), \label{eq:regPDE_b}\\
      \rho(0,x) = \rho_0(x),\quad \Phi(T,x) = -V(x).\label{eq:regPDE_c}
\end{numcases}
\end{subequations}
These coupled ODEs arose from adding regularizing Lagrangian terms $\beta \Delta_x$ to the ODEs given by the Benamou-Brenier formula. Here, $\Phi$ is a Kantorovich dual variable that has boundary condition $-V$ at time $T$. $\rho(T, x)$ is called the \emph{regularized Wasserstein proximal}. Using Hopf-Cole type transformations, \citet{li2023kernel} show the following closed-form integral representation for the regularized Wasserstein proximal
\begin{equation}\label{eq:rhoT}
    \rho(T,x) = \int_{\R^d} K(x,y) \rho_0(y)\, dy,
\end{equation}
\begin{equation}\label{eq:kernelDef}
    K(x,y) = \frac{\exp(-\frac{1}{2\beta} (V(x) + \frac{\|x-y\|^2}{2T}))}{\int_{\R^d} \exp(-\frac{1}{2\beta} (V(z) + \frac{\|z-y\|^2}{2T}))\, dz}.
\end{equation}

Observe that the normalizing constant in the kernel is given by a convolution between the potential $V$ and a heat kernel. We note that the integral formulation can be extended to $\rho(t,x)$ and $\Phi(t,x)$ for more general time $t\in [0,T]$, again given by a convolution with a heat kernel. 

We are interested in the solution of the Fokker-Planck equation
\begin{equation}\label{eq:Fokker_Planck}
    \frac{\partial \rho}{\partial t} = \nabla \cdot(\nabla V(x)\rho) + \beta \Delta \rho,\quad \rho(x,0) = \rho_0(x).
\end{equation}
We have the following relations between the Fokker-Planck equation and SDEs. More details can be found in \citet{jordan1998variational} and in references therein. The solution $\rho(t,x)$ of the Fokker-Planck equation is equal to the the density at time $t$ of the SDE
\begin{equation}\label{eq:FPSDE}
    d X(t) = -\nabla V(X(t)) d t + \sqrt{2\beta} dW(t),\quad X(0) = X_0,
\end{equation}
where $X_0$ is a random variable with density $\rho_0$. Under appropriate growth conditions of $V$ (such that the Gibbs measure is finite), the steady state of the Fokker-Planck equation (\ref{eq:Fokker_Planck}) is 
\begin{equation}
    \rho_\infty(x) \sim \exp(-\beta^{-1} V(x)).
\end{equation}

Moreover, the Fokker-Planck equation can be viewed as a Wasserstein gradient flow on the free energy \citep{otto2001geometry}. Thus, this steady state is the minimizer of the free energy functional $\gE$ over probability densities
\begin{equation}
    \mathcal{E}(\rho) = \int_{\R^d} \beta \rho \log \rho + V\rho\ dx.
\end{equation}

\subsection{Score Based Diffusion}
Instead of using a random particle formulation arising from a discretization of the SDE in \Cref{eq:FPSDE}, we can use a deterministic version, given knowledge of the density $\rho$ (which is intractable in practice). We now introduce the score based model, where particles are updated according to the gradient of the potential, and the score function $\nabla_x \log \rho(t,x)$. This formulation arises from Liouville's equation, which states the following \citep{liouville1838note,kardar2007statistical,tolman1979principles}.

\begin{proposition}[{\citealp{kubo1963stochastic}}]
    For an evolution under a density $\rho(t,x)$ given by a Hamiltonian $\gK$,
    \[\frac{d X}{d t} = \gK(t,X,\rho),\]
    the distribution function is constant along the trajectories. In particular, $\rho$ satisfies
    \begin{equation*}
        \frac{\partial \rho}{\partial t}+ \nabla \cdot (\rho \gK(t,x,\rho)) = 0.
    \end{equation*}
\end{proposition}

We can use Liouville's equation to derive an ODE for the Fokker-Planck dynamics (\ref{eq:Fokker_Planck}). Taking $\gK(t,x,\rho) = -\nabla V(x) - \beta \frac{\nabla \rho(t,x)}{\rho(t,x)}$ with $\nabla\log\rho=\frac{\nabla\rho}{\rho}$, we obtain the following ODE, with density $\rho(t,x)$ at time $t$ evolving as in the Fokker-Planck equation
\begin{equation}\label{eq:FPODE}
    \frac{d X}{d t} = -\nabla V(X) - \beta \nabla \log \rho(t,X).
\end{equation}
If we instead consider the regularized Fokker-Planck equation (\ref{eq:regPDE_a}), this approximates the Fokker-Planck dynamics (\ref{eq:Fokker_Planck}). Applying Liouville's equation with (\ref{eq:regPDE_a}) and $\gK(t,x,\rho) = \nabla \Phi(t,x) - \beta \nabla\log \rho$ for time $t \in [0,T]$, we obtain the following particle evolution ODE, whose density at time $t$ is equal to $\rho(t,x)$:
\begin{equation}\label{eq:LiouvillePDE}
    \frac{d X}{d t} = \nabla \Phi(t,X) - \beta \nabla \log \rho(t,X).
\end{equation}
The main difference between this regularized formulation and the non-regularized Fokker-Planck is that $V$ is replaced with the dual variable $\Phi$, and this evolution is only valid for $t \in [0,T]$. Both terms of \Cref{eq:LiouvillePDE} are problematic. Firstly, we do not have a closed form for $\Phi(t,x)$ for $t>T$ (though an integral formulation is available for $t<T$), and we only have the boundary condition $\Phi(T,x) = -V(x)$. Secondly, the score is not available. This work approximates the score using the score of the regularized Wasserstein proximal, which will be shown to have nice computational properties.

In the next section, we propose using the backwards Euler discretization method, utilizing only the boundary information for $\Phi$ and $\rho$, and thus only requiring $V$ and the regularized Wasserstein proximal $\rho_T = \rho(T,x)$. We will demonstrate that the combination of the kernel formulae \Cref{eqs:ClosedForms} and the backwards Euler discretization method, with an additional empirical approximation to the scores, result in a deterministic sampling method. In \Cref{sec:experiments}, we compare our proposed algorithm with ULA and MALA, starting with a mixing-time analysis for the special case of quadratic potentials, corresponding to the Ornstein-Uhlenbeck process. We additionally demonstrate convergent, \emph{structured} particle behavior, a variance reduction phenomenon, and improved performance on Bayesian logistic regression and Bayesian neural network training problems.




\section{Approximating the Score}\label{sec:Sec2}
In this section, we present the derivation and formulation of the proposed backwards regularized Wasserstein proximal (BRWP) scheme. Mixing time analysis is then given for the case where the target density is Gaussian, with closed-form updates for the mean and covariance. We characterize the discretization bias and demonstrate the convergence of the distribution to the regularized Wasserstein proximal of the target Gaussian distribution.

Our main goal is to approximately solve the ODE (\ref{eq:FPODE}) numerically for particles $X$, using approximations given by (\ref{eq:LiouvillePDE}). In this fashion, we are able to sample particles according to a distribution that evolves approximately according to the corresponding Fokker-Planck equation. The general idea is to consider the regularized Wasserstein proximal map as an approximation to the JKO scheme at each time step. We will demonstrate that the backwards Euler discretization of this approximate scheme is particularly amenable to computation. To begin, we consider the following four approximation steps. 

\textbf{Time approximation using the Wasserstein proximal.} For a small time $T$, the approximate Wasserstein proximal dynamics (\ref{eqs:regPDE}) approximates the Fokker-Planck dynamics (\ref{eq:Fokker_Planck}), where $\rho_0$ is replaced with $\rho(t,x)$. We thus approximate the Fokker-Planck dynamics by partitioning time into $[0,T],\, [T,2T],\, [2T,3T],...$ for $t\ge 0$, and approximating each $[kT, (k+1)T]$ using \Cref{eqs:regPDE}, and $\rho_{k,0}(x) = \rho(kT,x)$. We thus approximate the ODE (\ref{eq:FPODE}) with (\ref{eq:LiouvillePDE}) on each time partition. To compute this approximation, we can use the following techniques.

\textbf{Backwards discretization in time.} Since we only have particles at each iteration, analytic formulations of the densities $\rho_{k,0}$ and thus $\rho_{k,T}$ are unavailable. Instead of using kernel approximation methods or otherwise to approximate the score at time $t=kT$, we instead compute \emph{exactly} the score at time $t+T=(k+1)T$, conditional on $\rho_{k,0}$ being a sum of Dirac masses at the locations of the corresponding particles. This allows for implicit time steps of the Fokker-Planck equation, assuming knowledge of $\rho_{k,T}$. We can compute $\rho_{k,T}$ when $\rho_{k,0}$ is given by an empirical distribution as follows. 

\textbf{Computing using the kernel formulation.} For backwards Euler discretization, we need to know $\rho_{k,T}$ and $\Phi(T,\cdot)$ as evolved using \Cref{eqs:regPDE}. $\rho_{k,T}$ is given in \Cref{eq:rhoT} using a kernel convolution on $\rho_{k,0}$, and $\Phi(T, x) = -V(x)$ as defined in \Cref{eq:regPDE_c}. We note that while it is possible to perform a forward discretization on the $\Phi$ term by computing $\Phi(0,x)$, it is not possible on the $\rho$ term, as the score of a mixture of Dirac masses is undefined. Therefore, we apply a backwards Euler discretization of \Cref{eq:LiouvillePDE}.

\textbf{Convolution as sampling. } Observe the denominator in the convolution kernel given by \Cref{eq:kernelDef} takes the form of a Gaussian expectation. More precisely, this normalizing constant is given by a convolution of $\exp(-V)$ with a quadratic term. Using this trick similarly to \citet{osher2023hamilton}, we can compute the denominator of $K(x, y)$ by sampling from $z \sim \gN(y, 2T\beta)$. Moreover, noting the normalizing constants for the Gaussians cancel out, this form means that we can compute integrals using Gaussian expectations. These can be computed using Monte Carlo integration for distributions $f$ as follows.
\begin{equation}
    \int_{\R^d} f(y) K(x,y)\, dy = \frac{\E_{y \sim \gN(x, 2T\beta)}\left[f(y) \exp\left(- \frac{V(x)}{2\beta}\right)\right]}{\E_{z \sim \gN(x, 2T\beta)} \left[\exp\left(- \frac{V(z)}{2\beta}\right)\right]}.
\end{equation}

By combining these four approximation steps together, we obtain one step of the regularized Wasserstein proximal ODE \Cref{eq:LiouvillePDE}, discretized using the backwards Euler scheme. One discrete iteration with step-size $\eta>0$ can be written as
\begin{equation}\label{eq:BackwardsWProxStep}
\begin{split}
        X_{k+1} &= X_k + \eta\left[\nabla \Phi(T,X_k) - \beta \nabla \log \rho_{k,0}(T,X_k)\right]\\&= X_k - \eta \nabla V(X_k) - \eta \beta \nabla \log \rho_{k,T}(X_k).
\end{split}
\end{equation}
To turn this into a discrete update scheme, we consider at each step setting $\rho_{k,0}$ to be the empirical distribution of $X_k$, rather than $\rho_{k,0} = \rho_{k-1,T}$. If we have $N$ realizations of $X_k$ given by $\{\rvx_{k,i}\}_{i=1}^N$, we approximate $\rho_{k,0}$ using the empirical distribution,
\begin{equation*}
    \rho_{k,0} = \frac{1}{N}\sum_{i=1}^N \delta_{\rvx_{k,i}}.
\end{equation*}
Noting that $\nabla \log \rho_{k,T}(x) = \nabla \rho_{k,T}(x)/\rho_{k,T}(x)$, and using the closed-form expression $\rho_{k,T}(x) = \int \rho_{k,0}(y)K(x,y) dy$, we have the following expression for $\rho_{k,T}$ and the gradient  $\nabla \rho_{k,T}$ at a point $\rvx_i$, temporarily dropping the $k$ subscript:
\begin{subequations}\label{eqs:ClosedForms}
\begin{equation}\label{eq:rhoTClosedForm}
\begin{split}
    \rho_{k,T}(\rvx_i) &= \frac{1}{N}\sum_{j=1}^N K(\rvx_i, \rvx_j) 
    = \frac{1}{N}\sum_{j=1}^N \frac{\exp \left[-\frac{1}{2\beta}\left(V(\rvx_i) + \frac{\|\rvx_i - \rvx_j\|^2}{2T}\right)\right]}{\gZ(\rvx_j)}, \\
\end{split}
\end{equation}
\begin{equation}\label{eq:gradrhoTClosedForm}
\begin{split}
    \nabla \rho_{k,T}(\rvx_i) &= \frac{1}{N}\sum_{j=1}^N \frac{\left(-\frac{1}{2\beta}\left(\nabla V(\rvx_i) + \frac{\rvx_i - \rvx_j}{T}\right)\right)\exp \left[-\frac{1}{2\beta}\left(V(\rvx_i) + \frac{\|\rvx_i - \rvx_j\|^2}{2T}\right)\right]}{\gZ(\rvx_j)}, \\
\end{split}
\end{equation}
\begin{equation}\label{eq:normalizingConstant}
    \gZ(\rvx_j) \coloneqq \E_{z \sim \gN(\rvx_j, 2T\beta)} \left[\exp\left(- \frac{V(z)}{2\beta}\right)\right].
\end{equation}
\end{subequations}

This algorithm can be appropriately vectorized for parallelization. Indeed, as a kernel method, we need to compute the squared distances between all pairs of samples. This computational burden can be lessened by instead subsampling from the current samples to further approximate $\rho_{k,T}$. The Gaussian expectations can be done using Monte Carlo integration. The algorithm, consisting of \Cref{eq:BackwardsWProxStep,eqs:ClosedForms}, is detailed in full in \Cref{alg:BackwardWProxDiscrete}. Note that the loops can be vectorized to improve run-time by replacing the intermediate variables $\gZ, \gE, \gV$ with appropriately sized tensors.

\begin{algorithm}
\caption{Backwards regularized Wasserstein proximal (BRWP) scheme}\label{alg:BackwardWProxDiscrete}
\hspace*{\algorithmicindent} \textbf{Input:} Potential $V$, samples $(\rvx_{0,i})_{i=1}^N \sim \mu_0^{\otimes N}$, step-size $\eta>0$, regularization parameters $T,\beta>0$, Monte Carlo sample count $P$\\
\hspace*{\algorithmicindent} \textbf{Output:} Sequence of samples $(\rvx_{k,i})_{i=1}^N$ for $k=1,2,...$
\begin{algorithmic}[1]
\For{$k \in \mathbb{N}$} 
    \For{$i =1,...,N$}
        \State Sample $(\rvz_{k,i,p})_{p=1}^P \sim \gN(\rvx_{k,i}, 2\beta T I)$ \Comment{Sample for the expectation}
        \State $\gZ_{k,i} = \frac{1}{P} \sum_{p=1}^P \exp\left(-\frac{V(\rvz_{k,i,p})}{2\beta}\right)$ \Comment{Approximate $\gZ(\rvx_i)$ from (\ref{eq:normalizingConstant})}
    \EndFor
    \For{$i,j =1,...,N$}
    \State $\gE_{k,i,j} = \exp \left[-\frac{1}{2\beta}\left(V(\rvx_{k,i}) + \frac{\|\rvx_{k,i} - \rvx_{k,j}\|^2}{2T}\right)\right]$\Comment{Compute the numerator of (\ref{eq:rhoTClosedForm})}
    \State $\gV_{k,i,j} = -\frac{1}{2\beta}\left(\nabla V(\rvx_{k,i}) + \frac{\rvx_{k,i} - \rvx_{k,j}}{T}\right)$ \Comment{Compute the numerator of (\ref{eq:gradrhoTClosedForm})}
    \EndFor
    \For{$i =1,...,N$}
    \State $\nabla \log \rho_{k,T}(\rvx_{k,i}) = (\sum_j \gV_{k,i,j}\gE_{k,i,j}/\gZ_{k,j})/(\sum_j \gE_{k,i,j}/\gZ_{k,j})$ \Comment{Compute the score}
    \State $\rvx_{k+1,i}= \rvx_{k,i} - \eta \nabla V(\rvx_{k,i}) - \eta\beta\nabla \log \rho_{k,T}(\rvx_{k,i})$\Comment{Perform the update (\ref{eq:BackwardsWProxStep})}
    \EndFor
\EndFor
\end{algorithmic}
\end{algorithm}

A heuristic interpretation of the algorithm can be obtained by considering the score function as a weighted search direction. Indeed, $\log \rho_{k,T}$ is computed as a weighted sum of differences of $\gV_{k,i,j}$, which contains a $-(\rvx_{k,i} - \rvx_{k,j})$ term in its expression. Considering the update Step 12 in \Cref{alg:BackwardWProxDiscrete}, the sample particle $\rvx_{k,i}$ is repelled away from a weighted sum of all the particles. This is the mechanism through which this method achieves diffusion.

We note that Step 12 of \Cref{alg:BackwardWProxDiscrete} is a single gradient step on the free energy $V + \beta \log \rho_{k,T}$ applied to each of the particles $x_{k,i}$. This places a natural restriction on the step-size $\eta$, based on the Lipschitz constant of the free energy (at each time step). Informally, as $k\rightarrow \infty$, we should have that $\rho_{k,T} \rightarrow \rho_T$, where $\rho_T$ is the regularized Wasserstein proximal of the target distribution. However, the convergence analysis has to be delicate due to the changing density $\rho_{k,T}$ at each step.

\subsection{Closed Form Gaussian Evolution}
We begin our analysis with the simple case where $V$ is quadratic. Moreover, we find closed forms for the distribution at iteration $k$, given that the initial distribution is also Gaussian. Consider first the Ornstein-Uhlenbeck process without drift in one dimension, which is a special case of the Fokker-Planck equation. The governing SDE for a constant $a>0$ is as follows, where $W$ is a Wiener process \citep{karatzas1991brownian,gardiner1985handbook}:
\begin{equation}\label{eq:OUProcess}
    d X = - a X d t + \sqrt{2\beta} d W.
\end{equation}
This can be seen as taking the potential to be $V(x) = a x^2/2$. The true solution for initialization $X_0$ is given by
\begin{equation}
    X_t = X_0 e^{-at} + \frac{\sqrt{2\beta}}{\sqrt{2a}} W_{1 - e^{-2at}}.
\end{equation} 

If $X_0$ is initially normally distributed with mean $\mu_0$, variance $\sigma_0^2$, then the distribution at $X_t$ will also be normally distributed, with means $\mu_t$ and variance $\sigma^2_t$ given by
\begin{equation*}
    \mu_t = \mu_0 e^{-at},\quad \sigma_t^2 = \sigma_0^2e^{-2at} + \frac{\beta}{a} (1-e^{-2at}).
\end{equation*}
The steady state of the flow is Gaussian with mean and variance
\begin{equation*}
    \mu_\infty = 0,\quad \sigma^2_\infty = \frac{\beta}{a}.
\end{equation*}

To discretize this flow, we consider two competing methods, ULA and MALA. We can compute the analytic solutions with quadratic potential $V = ax^2/2$ and Gaussian distributed initializations $X_0 \sim \gN(\mu_0, \sigma_0^2)$. 

\textbf{ULA.} For a step-size $\eta<a^{-1}$, ULA consists of an explicit Euler-Maruyama discretization of \Cref{eq:OUProcess}:
\begin{equation*}
    X_{k+1} = (1-a\eta) X_k + \sqrt{2\beta\eta}Z_k,
\end{equation*}
where $(Z_k)_{t=0}^\infty$ are i.i.d standard Gaussians. Therefore, $X_t$ are also Gaussian, with mean and variance satisfying the recurrence relations
\begin{equation*}
    \mu_{k+1} = (1-a\eta) \mu_k,\quad \sigma_{k+1}^2 = (1-a\eta)^2 \sigma_k^2 + 2\beta \eta.
\end{equation*}
Solving the recurrence relations gives the closed form solutions
\begin{equation*}
    \mu_k = (1-a\eta)^k \mu_0,\quad \sigma_k^2 = (1-a \eta)^{2k} \sigma_0^2 + 2 \beta\eta \sum_{j=0}^{k-1} (1-a\eta)^{2j}.
\end{equation*}
Observe that the variance is biased due to the explicit discretization \citep{wibisono2018sampling}:
\begin{equation*}
    \lim_{k\rightarrow \infty} \sigma_k^2 = \frac{2 \beta}{(2-a\eta) a} > \frac{\beta}{a} = \sigma_\infty^2.
\end{equation*}

\textbf{MALA.} The Metropolis-adjusted Langevin algorithm introduces an additional Metropolis-Hastings acceptance step after ULA \citep{dwivedi2018log,roberts1996exponential}. The MALA update is as follows in the case where $\beta = 1$.
\begin{gather*}
    \tilde X_{k+1} = (1-\eta \nabla V) X_k + \sqrt{2\eta}Z_k; \\
    \alpha_{k+1} = \min \left\{1, \frac{\exp\left( -V(\tilde X_{k+1}) - \|X_k - \tilde{X}_{k+1} + \eta \nabla V(\tilde{X}_{k+1})\|^2/4\eta \right)}{\exp\left( -V(X_k) - \|\tilde{X}_{k+1} - X_k + \eta \nabla V(X_k)\|^2/4\eta\right)}\right\};\\
    X_{k+1} = \begin{cases}
        \tilde{X}_{k+1}, & \text{with probability } \alpha_{k+1};\\
        X_{k}, & \text{with probability } 1-\alpha_{k+1}.\\
    \end{cases}
\end{gather*}
In the case that $\beta \ne 1$, we can perform a change of variables by considering step-size $ \beta \tilde\eta$ and potential $V/\beta$. Then the MALA scheme will have modified acceptance probabilities of the form 
\begin{gather*}
    \tilde X_{k+1} = (1-\eta \nabla V) X_k + \sqrt{2\beta\eta}Z_k; \\
    \alpha_{k+1} = \min \left\{1, \frac{\exp\left( -V(\tilde X_{k+1})/\beta - \|X_k - \tilde{X}_{k+1} + \eta \nabla V(\tilde{X}_{k+1})\|^2/(4\beta\eta) \right)}{\exp\left( -V(X_k)/\beta - \|\tilde{X}_{k+1} - X_k + \eta \nabla V(X_k)\|^2/(4\beta\eta)\right)}\right\};\\
    X_{k+1} = \begin{cases}
        \tilde{X}_{k+1}, & \text{with probability } \alpha_{k+1};\\
        X_{k}, & \text{with probability } 1-\alpha_{k+1}.\\
    \end{cases}
\end{gather*}
We note that $X_k$ does not follow a Gaussian distribution due to this acceptance step. MALA is unbiased, and converges in distribution to the target Gaussian $\gN(\mu_\infty, \sigma_\infty^2)$.

\textbf{BRWP.}  Assuming $X_k \sim \gN(\mu_k, \sigma_k^2)$, we can compute the closed form of $\rho_{k,T}(x)$ with initial condition $\rho_{k,0} \sim \gN(\mu_k, \sigma_k^2)$ using the kernel formulation. A full derivation can be found in \Cref{app:updateGaussian}. The approximate Wasserstein proximal $\rho_{k,T} \sim \gN(\tilde\mu_{k+1}, \tilde\sigma_{k+1}^2)$ is Gaussian, with mean and variance
\begin{equation}\label{eq:1dGaussianrhoT}
    \tilde\mu_{k+1}
    = \frac{\mu_k}{1+aT},\quad \tilde\sigma_{k+1}^2 = \frac{\sigma_k^2}{(1+aT)^2} + \frac{2\beta T}{1+aT}.
\end{equation}
Applying the discrete backwards iteration given in \Cref{eq:BackwardsWProxStep} with this closed form for $\rho_{k,T}$, we have
\begin{align*}
    X_{k+1} &= (1-a\eta) X_k - \eta \beta \nabla\log\rho_{k,T}(X_k) \\
    &= (1-a \eta)X_k + \eta \beta \frac{X_k - \tilde\mu_{k+1}}{\tilde\sigma_{k+1}^2} \\
    &= (1-a\eta + \frac{\eta\beta}{\tilde{\sigma}^2_{k+1}})X_k - \frac{\eta \beta \tilde{\mu}_{k+1}}{\tilde{\sigma}_{k+1}^2} \\
    &= \left(1-a\eta + \frac{\eta\beta(1+aT)^2}{\sigma_k^2+2\beta T(1+aT)}\right)X_k - \frac{\eta \beta \mu_k (1+aT)}{\sigma_k^2 + 2\beta T(1+aT)}.
\end{align*}
Therefore, $X_{k+1}$ is Gaussian, with mean and variance satisfying the recurrence relations
\begin{subequations}\label{eqs:WPKernelDiscrete}
\begin{align}
    \mu_{k+1} &= \left(1-a\eta + \frac{\eta\beta(1+aT)^2}{\sigma_k^2+2\beta T(1+aT)}\right)\mu_k - \frac{\eta \beta \mu_k (1+aT)}{\sigma_k^2 + 2\beta T(1+aT)} \notag\\
    &= \left(1-a\eta + \frac{\eta\beta aT(1+aT)}{\sigma_k^2+2\beta T(1+aT)}\right)\mu_k,\label{eq:WPKernelMeanRecRel}\\
    \sigma_{k+1}^2 &= \left(1-a\eta + \frac{\eta\beta(1+aT)^2}{\sigma_k^2+2\beta T(1+aT)}\right)^2 \sigma_k^2. \label{eq:WPKernelVarRecRel}
\end{align}
\end{subequations}

We can compute the steady states of \Cref{eqs:WPKernelDiscrete} by setting the front term in \Cref{eq:WPKernelVarRecRel} to 1. We can do this by assuming $a\eta<2$, which is a sufficient condition for stability of the recurrence. This results in 
\begin{gather*}
    \left(1-a\eta + \frac{\eta\beta(1+aT)^2}{\sigma_\infty^2+2\beta T(1+aT)}\right) = 1 \\ 
    \Longrightarrow \sigma_\infty^2 = \frac{\beta}{a}(1-a^2T^2) \text{ if } aT<1,\, \sigma_\infty^2 = 0 \text{ otherwise}.
\end{gather*}
We observe that the bias is different to ULA due to the backwards discretization. Indeed, the bias of ULA results in a variance that is larger than the target variance. On the other hand, the bias for BRWP results in a variance that is smaller than the target variance, and moreover does not depend on the step-size $\eta$.

\begin{figure}
    \centering
    \subfloat[\centering Mean]{{\includegraphics[height=5cm]{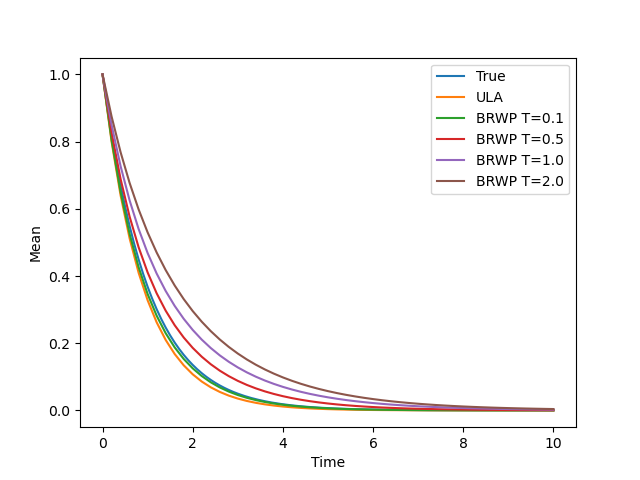} }}%
    \subfloat[\centering Variance]{{\includegraphics[height=5cm]{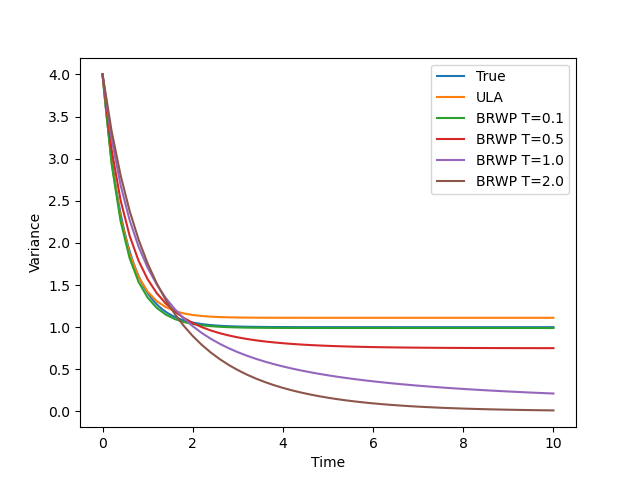} }}
    
    \caption{Evolution of the analytic solutions for ULA and kernel formula, with initialization $\gN(0,4)$ and target distribution $\gN(0,1)$. The parameters of the OU flow are $a=\beta=1$, discretized with step-size $\eta = 0.25$. The larger $T$ is, the smaller the stationary variance is. We observe that when $T \ge 1$, the stationary distribution under the Wasserstein proximal flow is degenerate.}
    \label{fig:gaussian_analytic}
\end{figure}

\subsection{Multi-dimensional Gaussian}
With some care, we can extend the analysis of our previous section to the multi-dimensional case, and again obtain closed form expressions for the mean and variance at iteration $k$. Suppose now that we are working in $\R^d$, and our $V$ takes the form for a zero-mean Gaussian with (symmetric positive-definite) covariance $\Sigma^{-1}$
\begin{equation}
    V(x) = \frac{1}{2}x^\top \Sigma^{-1} x.
\end{equation}

As before, we can obtain a closed form for the approximate Wasserstein proximal, and the derivation can be found in \Cref{app:updateGaussian}. We have that $\rho_{k,T} \sim \gN(\tilde\mu_{k+1}, \tilde\Sigma_{k+1})$, with mean and covariance
\begin{subequations}\label{eqs:nDGaussianrhoT}
\begin{equation}
    \tilde{\Sigma}_{k+1}^{-1} = \left(2\beta T (I + T\Sigma^{-1})^{-1} + (I + T\Sigma^{-1})^{-1}\Sigma_k (I + T\Sigma^{-1})^{-1} \right)^{-1},
\end{equation}
\begin{equation}
    \tilde{\mu}_{k+1} = (I + T\Sigma^{-1})^{-1} \mu_k.
\end{equation}
\end{subequations}

Applying the discrete backwards iteration \Cref{eq:BackwardsWProxStep},
\begin{align*}
    X_{k+1} &= X_k - \eta\nabla V( X_k ) - \eta \beta \nabla \log \rho_{k,T}(X_k) \\
    &= (I - \eta \Sigma^{-1})X_k + \eta \beta \tilde\Sigma_{k+1}^{-1}(X_k - \tilde{\mu}_{k+1}) \\
    &= (I - \eta \Sigma^{-1} + \eta\beta\tilde\Sigma_{k+1}^{-1})X_k - \eta\beta\tilde\Sigma_{k+1}^{-1} \tilde\mu_{k+1}.
\end{align*}
Since $X_k$ is Gaussian and affine transformations of Gaussian distributions are Gaussian, we can obtain the following recurrence relations for the parameters of $X_{k+1}$.
\begin{proposition}
$X_{k+1}$ is Gaussian with mean $\mu_{k+1}$ and covariance $\Sigma_{k+1}$ given by
\begin{subequations}
\begin{align}
    \mu_{k+1} &= (I - \eta\Sigma^{-1})\mu_k + (\eta\beta\tilde{\Sigma}_{k+1}^{-1})(\mu_k - \tilde{\mu}_{k+1}) \notag\\
    &= \left(I - \eta\Sigma^{-1} + \eta\beta \left(2\beta T I + \Sigma_k (I + T\Sigma^{-1})^{-1} \right)^{-1}(T\Sigma^{-1})\right)\mu_k, 
\end{align}
\begin{align}\label{eq:MultiDimGaussianVar}
    \Sigma_{k+1} &= (I - \eta\Sigma^{-1} + \eta\beta\tilde\Sigma_{k+1}^{-1}) \Sigma_{k} (I - \eta\Sigma^{-1} + \eta\beta\tilde\Sigma_{k+1}^{-1})^{\top}. 
\end{align}
\end{subequations}
\end{proposition}
We obtain the covariance of the stationary distribution by setting $I - \eta\Sigma^{-1} + \eta\beta\tilde\Sigma_{k+1}^{-1} = I$, yielding 
\begin{equation}
    \Sigma_\infty = \beta (I - T\Sigma^{-1}) \Sigma (I + T\Sigma^{-1}).
\end{equation}

\subsubsection{Mixing Time: Gaussian}
We first consider the case where the mean of the initialization is zero, and further that the covariance of the initialization commutes with the covariance of the target distribution. Observe that if $\Sigma$ commutes with $\Sigma_{k}$, then $\Sigma$ commutes with $\tilde\Sigma_{k+1}$ and hence with $\Sigma_{k+1}$. Therefore, without loss of generality, we can work in a simultaneously diagonal basis for $\Sigma$, and all $\Sigma_k$ and $\tilde\Sigma_k$. We show linear convergence of the eigenvalues of the covariance matrix to those of the stationary distribution, and give the rate of convergence in terms of $T$.

We aim to bound the mixing time of the Gaussians under the approximate Wasserstein iterations, defined as follows. This is a measure of how quickly a sequence of distributions converges to a target distribution. 

\begin{definition}[Mixing time]
    The \emph{total variation} between two probability distributions over a measurable space $(\Omega, \mathcal{F})$ is 
    \begin{equation*}
        \TV(\gP, \gQ) = \sup_{A \in \mathcal{F}} |\gP(A) - \gQ(A)|.
    \end{equation*}
    For a operator $\gT_p$ on the space of probability distributions, assume that the chain $\gT_p^k(\mu_0) \rightarrow \Pi$ as $k \rightarrow \infty$ for some probability distribution $\Pi$. The \emph{$\delta$-mixing time} with $\delta \in (0,1)$ and initial distribution $\mu_0$ is 
    \begin{equation*}
        t_{\text{mix}}(\delta;\mu_0) = \min \left\{k \mid \TV(\gT_p^k(\mu_0), \Pi) \le \delta\right\}.
    \end{equation*}
\end{definition}
We have the following theorem upper-bounding the total variation between two Gaussians with the same mean. This means that we can control the total variation between two Gaussians with the difference between the covariance matrices.

\begin{theorem}[{\citealp[Thm. 1.1]{devroye2018total}}]\label{thm:GaussianTV}
    Let $\mu \in \R^d$, $\Sigma_1,\Sigma_2$ be two positive-definite $d\times d$ covariance matrices, and $\lambda_1,...,\lambda_d$ denote the eigenvalues of $\Sigma_1^{-1} \Sigma_2 - I$. Then the total variation satisfies
    \begin{equation}
        \TV(\gN(\mu, \Sigma_1), \gN(\mu, \Sigma_2)) \le \frac{3}{2} \min\left\{1, \sqrt{\sum_{i=1}^d \lambda_i^2}\right\}.
    \end{equation}
\end{theorem}

We firstly assume that $\mu_0 = \mathbf{0}$. Observe that this means that $\mu_k = \mathbf{0}$ for all $k$. Suppose further that $\Sigma_0$ commutes with $\Sigma$, for example if $\Sigma_0 = c_0 I$ for some $c_0>0$. Under this assumption, we can simultaneously diagonalize $\Sigma_0$ and $\Sigma$ (and indeed, $\Sigma_k$ as well for all $k$). Without loss of generality, let us work in an orthonormal eigenbasis, so that $\Sigma = \diag(\xi^{(1)}, ..., \xi^{(d)})$ and $\Sigma_k = \diag(\tau^{(1)}_k, ..., \tau^{(d)}_k)$ are all diagonal. The following theorem states that the covariance under the BRWP iterations converge linearly to the stationary distribution.

\begin{theorem}[Mixing time for multi-dimensional Gaussians with same initialization mean]\label{thm:MixingTime}
    Consider the regularized Wasserstein proximal scheme applied to the zero-mean Ornstein-Uhlenbeck process in $d$ dimensions
    \[dX = -\nabla V(X) dt + \sqrt{2\beta} dW,\quad V(x) = \frac{1}{2} x^\top \Sigma^{-1} x,\]
    where $\Sigma = \diag(\xi^{(1)}, ..., \xi^{(d)})$ is positive definite. Let $T>0, \eta>0$ be such that $T < \min \{\xi^{(i)} \mid i = 1,..., d\}$, and $\eta \xi^{(i) -1} \le 1/\Delta(\xi^{(i)}, T)$ for all $i$, where 
    \begin{equation*}
        \Delta(\xi^{(i)}, T) = \frac{1}{2}\left(\sqrt{\frac{\xi^{(i)}+T}{2T}}+1\right).
    \end{equation*}
    The stationary distribution $\Pi$ of the discrete scheme \Cref{eq:BackwardsWProxStep} is given by
    \begin{equation}
        \Pi \sim \gN(0, \Sigma_\infty),\, \Sigma_\infty = \diag( \tau_\infty^{(i)} \mid i=1,...,d)
    \end{equation}
    \begin{equation}
        \tau_\infty^{(i)} = \beta\xi^{(i)} (1-T^2 \xi^{(i)-2}), i=1,...,d.
    \end{equation}
    Suppose the chain is initialized with 
    \begin{equation*}
        X_0 \sim \gN(0, \Sigma_0),\, \Sigma_0 = \diag(\tau_0^{(i)} \mid i = 1,...,d),
    \end{equation*}
    with $\tau_0^{(i)} > 0$ for $i=1,...,d$. If $(X_k)_{k\ge0}$ evolves under the BRWP scheme \Cref{eq:BackwardsWProxStep}, we have the following closed-form for the distributions of $X_k$:
    \begin{equation*}
        X_k \sim \gN(0, \Sigma_k),\, \Sigma_k = \diag(\tau_k^{(i)} \mid i = 1,...,d),
    \end{equation*}
    \begin{equation}
        \tau_{k+1}^{(i)} = \left(1 - \eta \xi^{(i)-1} + \frac{\eta\beta (1+T\xi^{(i)-1})^2}{\tau_k^{(i)}+2\beta T(1+T\xi^{(i)-1})}\right)^2\tau_k^{(i)},\quad i=1,...,d. \cc
    \end{equation}
    In particular, the eigenvalues of $\Sigma_k \Sigma_\infty^{-1} - I$ are given by 
    \begin{equation}
        \lambda_k^{(i)} = [\tau_k^{(i)} - \tau_\infty^{(i)}]/\tau_\infty^{(i)},
    \end{equation}
    which converge linearly to 0 with rate of convergence $[1-\eta\xi^{(i)-1}(\xi-T)/(\xi+T)] \in (0,1)$. Moreover, the total variation satisfies
    \begin{equation}
        \TV(\mu(X_k), \Pi_T) \le \frac{3}{2} \min \left\{1, \sqrt{\sum_{i=1}^d (\lambda_k^{(i)})^2}\right\} \le \frac{3}{2}C\sqrt{d}c^k,
    \end{equation}
    where $C = C(\Sigma_0,\Sigma, T)>0$ is the root mean squared of the initial eigenvalues of $\Sigma_0 \Sigma_T^{-1} - I$, and $c = c(\Sigma_0, \Sigma, T,\beta, \eta) \in (0,1)$ is the largest rate of convergence. Therefore the mixing time satisfies
    \begin{equation}
        t_{\text{mix}}(\delta, \mu(X_0)) = \mathcal{O}(\log (C \sqrt{d}/\delta)/\log(c)).
    \end{equation}
\end{theorem}

\begin{remark}
    The conditions mean that we want $T$ to be small to reduce the asymptotic bias, but also sufficiently large so that we can take a large step-size $\eta$.
\end{remark}

\begin{proof}[Sketch proof]
    Without loss of generality, we can assume all covariance matrices are diagonal. Using the closed-form update \Cref{eq:WPKernelVarRecRel} for the variance, we derive a recurrence relation for the eigenvalues of the covariance matrices. This recurrence relation converges to the stationary distribution linearly, provided that the step-size is chosen to be sufficiently small. A full proof can be found in \Cref{app:evalueGaussians}.
\end{proof}

For fixed $T<\min_i \xi^{(i)}$, let us compute $c$ in terms of the condition number $\kappa = L/m$ for initializations $\Sigma_0 = L^{-1}(1-L^{-2}T^2)^{-1} I$, where $m$ and $L$ are the smallest and largest eigenvalues of $\nabla^2 V = \Sigma^{-1}$ respectively. Note that $L(1-L^{-2} T^2) = \lambda_{\max}(\Sigma_\infty)$ is the Lipschitz constant of the log-Hessian of the stationary density. Without loss of generality, let the eigenvalues of $\Sigma$ be sorted in descending order, so that $L^{-1} = \xi^{(d)}, m^{-1} = \xi^{(1)}$. In this case, we have that $\omega^{(i)}(\gamma_k^{(i)}) \ge 0$ for each $i=1,...,d$ and all $k \ge 0$. Thus 
    \begin{equation*}
        \delta^{(i)} = \min(\omega^{(i)}(0),\omega^{(i)}(\gamma_0^{(i)} )) \ge \min\left(\frac{2(\xi^{(i)}-T)}{\xi^{(i)}+T}, 1\right) \ge \min\left(\frac{2(\xi^{(d)}-T)}{\xi^{(d)}+T}, 1\right) .
    \end{equation*} Let $\eta = \min_i \{\xi^{(i)}/\Delta^{(i)}\} =\xi^{(d)}/\Delta^{(d)} $ be the maximum allowed step-size. Then,
    \begin{equation}
    \begin{split}
        c &= \max_i \left\{1-\eta \xi^{(i)-1} \delta^{(i)}\right\} \\
        &= \max_i \left\{1-\eta \xi^{(i)-1} \delta^{(i)}\right\} \\
        & \le 1 - (\xi^{(d)}/\Delta^{(d)})\xi^{(1)-1} \min\left(\frac{2(\xi^{(d)}-T)}{\xi^{(d)}+T}, 1\right) \\
        & = 1 - \kappa^{-1}\min\left(\frac{2(\xi^{(d)}-T)}{\xi^{(d)}+T}, 1\right)/\left(\frac{1}{2} (\sqrt{(\xi^{(d)} + T)/(2T)} + 1)\right).
    \end{split}
    \end{equation}
    The choice of $T$ here is not particularly important as long as it is smaller than $\xi^{(d)}$. Taking $T = \xi^{(d)}/3$, we get
    \begin{equation*}
        c \le 1-1/\left(\frac{\kappa}{2} (\sqrt{3\kappa/2 + 1/2} + 1)\right),
    \end{equation*}
    so that $-1/\log c  = \mathcal{O}(\kappa^{3/2})$. Further note that for this choice of $T$, $C = 3(\kappa-1)/2$. We get the following.

\begin{corollary}
    For initialization $X_0 \sim \gN(0, L^{-1}(1-L^{-2}T^2)^{-1} I)$, where $m I \preceq \nabla^2 V \preceq L I$, and $\kappa = L/m$. Let $T = \xi^{(d)}/3$ and $\eta = \xi^{(1)}/\Delta^{(1)}$. The worst-case mixing time satisfies
    \begin{equation}
        t_{\text{mix}}(\delta, \mu(X_0)) = \mathcal{O}(\kappa^{3/2} \log (\kappa \sqrt{d}/\delta)).
    \end{equation}
\end{corollary}
As a comparison, we have that the mixing times with initialization $\gN(0, L^{-1} I)$ for ULA is $\mathcal{O}((d^3+d\log^2(1/\delta))\kappa^2\delta^{-2})$, and the mixing time for MALA is $\mathcal{O}(d^2\kappa \log(\kappa/\delta))$ \citep{dalalyan2017theoretical,dwivedi2018log}. We note that in the analytic case, our mixing time has a small dependence on the dimension, which comes only from translating convergence of the eigenvalues to convergence of the total variation distance. 

\subsection{Non-commuting Gaussian}
We now turn our attention to the case where the initialization $\Sigma_0$ does not commute with the target covariance $\Sigma$. We stay in the zero mean case. By considering the continuous limit of the BRWP updates, we show that the regularized Wasserstein proximals of the covariances converges to the regularized Wasserstein proximal of the target covariance in terms of Frobenius distance, $\|\ttS_t^{-1} - \ttS_\infty^{-1}\|_F^2 \rightarrow 0$. For ease of notation, let us first define $K \coloneqq I + T\Sigma^{-1}$. Note that $K$ commutes with $\Sigma$ and moreover is positive definite. We first recall the identities:
\begin{subequations}\label{eqs:NonCommuteGaussianDiscrete}
\begin{gather}
    \ttS_{k+1} = K^{-1} \Sigma_{k} K^{-1} + 2\beta T K^{-1}, \\
    \Sigma_k = K \ttS_{k+1} K - 2\beta T K, \\
    \Sigma_{k+1} = (I - \eta \Sigma^{-1} + \eta\beta \ttS_{k+1}^{-1}) \Sigma_k (I - \eta \Sigma^{-1} + \eta\beta \ttS_{k+1}^{-1}).
\end{gather}
\end{subequations}

Moreover, recall that the regularized Wasserstein proximal of the target distribution is $\ttS_\infty = \beta \Sigma$. Reformulating in terms of $\ttS$,

\begin{align*}
    \ttS_{k+2} &= K^{-1} \Sigma_{k+1} K^{-1} + 2\beta TK^{-1} \\
    &= 2\beta T K^{-1} + K^{-1}\left[I - \eta\beta(\ttS_\infty^{-1} - \ttS_{k+1}^{-1})\right]\Sigma_k\left[I - \eta\beta(\ttS_\infty^{-1} - \ttS_{k+1}^{-1})\right]K^{-1} \\
    &= 2\beta T K^{-1} + K^{-1} \Sigma_k k^{-1} \\
    & \qquad - K^{-1} \eta\beta (\ttS_\infty^{-1} - \ttS_{k+1}^{-1}) \Sigma_k K^{-1} \\
    & \qquad - K^{-1} \Sigma_k \eta\beta (\ttS_\infty^{-1} - \ttS_{k+1}^{-1}) K^{-1} \\
    & \qquad + K^{-1} \eta\beta (\ttS_\infty^{-1} - \ttS_{k+1}^{-1}) \Sigma_k \eta\beta (\ttS_\infty^{-1} - \ttS_{k+1}^{-1}) K^{-1} \\
    &= \ttS_{k+1} - \left[\eta\beta K^{-1} (\ttS_\infty^{-1} - \ttS_{k+1}^{-1}) K \right]\left[K^{-1} \Sigma_k K^{-1} \right] \\
    & \qquad - \left[K^{-1} \Sigma_k K^{-1} \right] \left[\eta\beta K (\ttS_\infty^{-1} - \ttS_{k+1}^{-1}) K^{-1} \right] \\
    & \qquad + \left[\eta\beta K^{-1} (\ttS_\infty^{-1} - \ttS_{k+1}^{-1}) K \right]\left[K^{-1} \Sigma_k K^{-1} \right]\left[\eta\beta K (\ttS_\infty^{-1} - \ttS_{k+1}^{-1}) K^{-1} \right].
\end{align*}

We see that this is a discretization of the continuous case by discarding the higher order $\eta$ terms, defined as follows,
\begin{subequations}\label{eqs:ctsEvo}
    \begin{equation}
    d\ttS_t/dt = -\beta K^{-1}\left[(\ttS_\infty^{-1} - \ttS_t^{-1}) \Sigma_t + \Sigma_t(\ttS_\infty^{-1} - \ttS_t^{-1}) \right] K^{-1},
    \end{equation}
    \begin{equation}
        \Sigma_t = K\ttS_{t}K - 2\beta T K.
    \end{equation}
\end{subequations}

The above discrete iteration \ref{eqs:NonCommuteGaussianDiscrete} for $\ttS_k$ is a discretization of the above ODE \ref{eqs:ctsEvo} in the limit as $\eta \rightarrow 0$. We find that the Frobenius norm $\|\ttS_\infty^{-1} - \ttS_t^{-1}\|_F^2$ is a Lyapunov function for the ODE formulation of the BRWP scheme. 

\begin{proposition}\label{prop:frobeniusLyapunov}
    The squared Frobenius norm $\|\ttS_\infty^{-1} - \ttS_t^{-1}\|_F^2$ is a Lyapunov function for the continuous limit of the BRWP scheme. Moreover, it converges linearly to zero.
\end{proposition}
\begin{proof}[Sketch proof]
    The time derivative of the Frobenius norm is given by the trace of the product of a positive definite matrix, and a matrix whose spectrum lies in the positive half line. Using the generalized H\"older's inequality for matrices, we upper bound the time derivative by a negative quantity that is proportional to the squared eigenvalues of $\ttS_\infty^{-1} - \ttS_t^{-1}$. A full proof can be found in \Cref{app:convergenceGaussian}.
\end{proof}

In this section, we used the closed form solution for the BRWP scheme to compute the evolution for the Ornstein-Uhlenbeck process. Expressions for the stationary solution and iterations were computed, and linear convergence to the stationary solutions were shown, with better dimension dependence on the mixing time compared to ULA and MALA.  


\section{Experiments}\label{sec:experiments}
For numerical experiments, we compare our method against ULA and MALA, using the experiments in \citet{dwivedi2018log,wang2022accelerated}. In particular, we consider target densities from an ill-conditioned Gaussian, a Gaussian mixture, a bimodal toy distribution, and additional experiments in Bayesian logistic regression and Bayesian neural network training. We will use this to demonstrate convergence to the (approximate) stationary distribution, as well as effectiveness without the requirement of pre-conditioning. Moreover, we will demonstrate the effect of using an ODE to model the particle movement instead of discretizing an SDE, in that the samples do not evolve significantly after some time. We compute ULA and MALA using the algorithms defined in \Cref{sec:Sec2}, and fix $\beta = 1$ for simplicity. The code for the experiments is publicly available on GitHub\footnote{\url{https://github.com/hyt35/WassersteinProxSampling}}.



\subsection{Ill-Conditioned Gaussian}\label{sec:ExperimentGaussian}
We first consider the case of a 2-dimensional and 5-dimensional Gaussian, with mean zero and diagonal covariance with eigenvalues evenly spaced from 10 to 1. The corresponding potential $V = x^\top \Sigma^{-1} x/2$ has Lipschitz constant $L=1$ and strong convexity parameter $m=0.1$. We consider the step-sizes to be $\eta = 0.1$ for ULA, MALA and the proposed BRWP scheme. For the BRWP scheme, we consider the choices $T=0.05, 0.1, 0.25, 0.5$. Note that the theory restricts $T<\lambda_{\min}(\Sigma) = 1$, so these choices of $T$ are valid and do not produce degenerate Gaussians for a closed form evolution. The number of Monte-Carlo samples used for computing the normalizing constant was set to $P=10$. We present three experiments, with dimension $d$ and number of samples $N$ as $(d,N) = (2,1000), (5,1000), (5,200)$, with samples initialized as $\gN(0, L^{-1}I) = \gN(0,I)$. We present two main findings, that we demonstrate further in following experiments.

\begin{figure}[]
    \centering
    \subfloat[\centering ULA]{{\includegraphics[height=2.5cm]{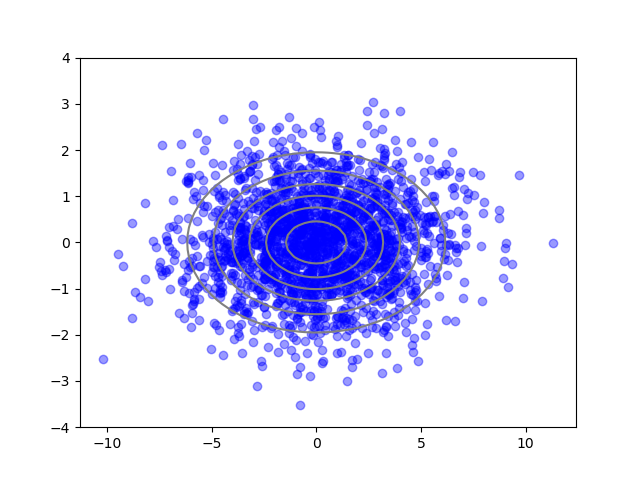}}}
    \subfloat[\centering MALA]{{\includegraphics[height=2.5cm]{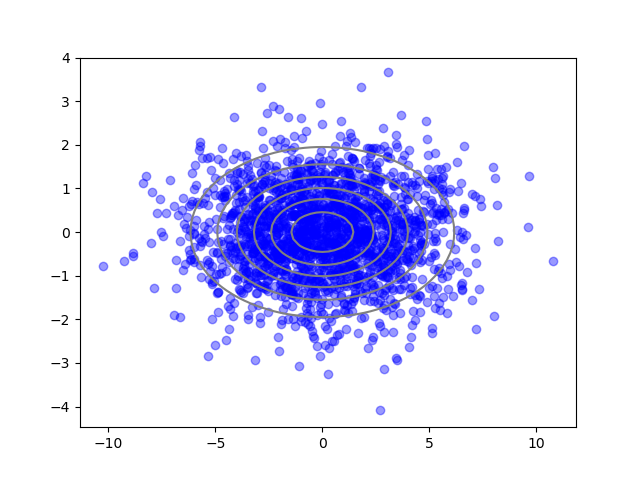}}}
    \subfloat[\centering BRWP $T=0.05$]{{\includegraphics[height=2.5cm]{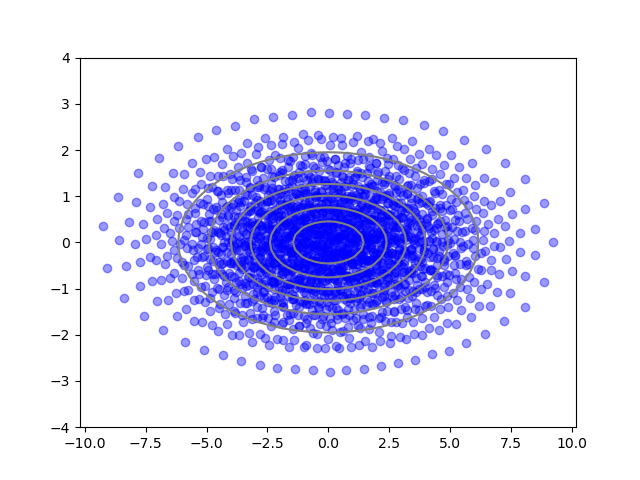}}}
    \subfloat[\centering BRWP $T=0.25$]{{\includegraphics[height=2.5cm]{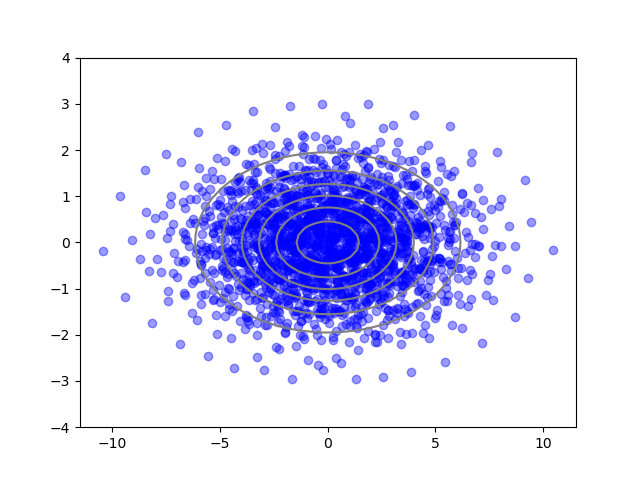}}}
    
    \caption{1000 particles after 200 iterations of ULA, MALA and BRWP with $T=0.05, 0.25$, applied to a two-dimensional Gaussian with condition number $\kappa=10$. We observe the samples for BRWP organize themselves into rings, clearer for $T=0.05$ than for $T=0.25$. This is as opposed to the randomness of ULA and MALA.}
    \label{fig:gaussian_2_1000}
\end{figure}

\textbf{Samples are structured.} \Cref{fig:gaussian_2_1000} demonstrates the effect of the deterministic sampling. In two dimensions, we observe a clear ellipsoidal structure that is traced out by the iterates, closely matched by the level set contours of the density $\exp(-V)$. This appears to be a consequence of both determinism as well as evolving an empirical approximation to the density at each iteration.

\begin{figure}[]
    \centering
    \subfloat[\centering BRWP $T=0.05$]{{\includegraphics[height=2.5cm]{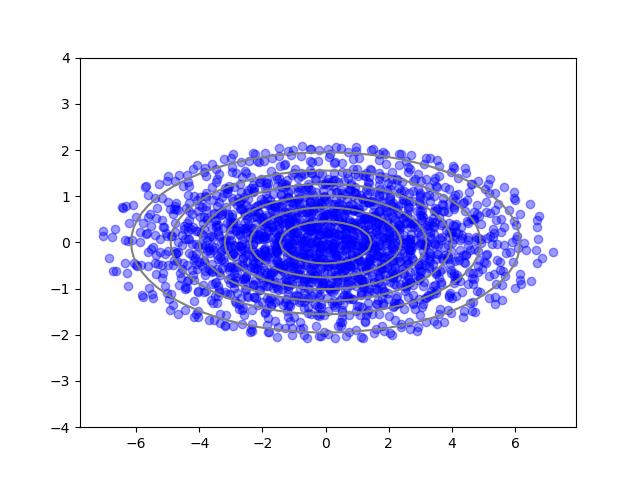}}}
    \subfloat[\centering BRWP $T=0.25$]{{\includegraphics[height=2.5cm]{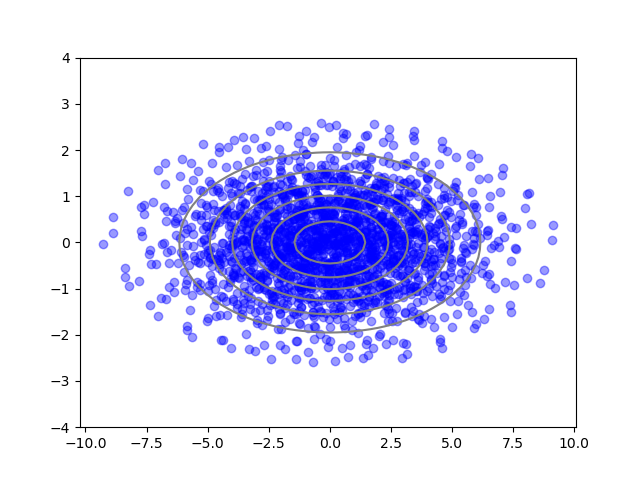}}}
    \subfloat[\centering BRWP $T=0.5$]{{\includegraphics[height=2.5cm]{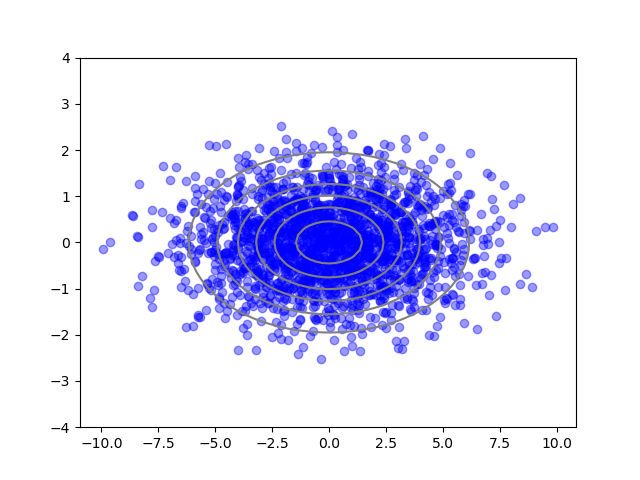}}}
    \subfloat[\centering BRWP $T=0.999$]{{\includegraphics[height=2.5cm]{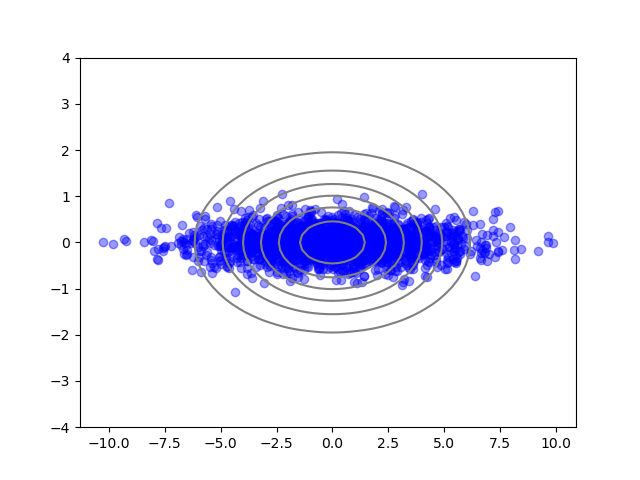}}}
    
    \subfloat[\centering BRWP $T=0.05$]{{\includegraphics[height=2.5cm]{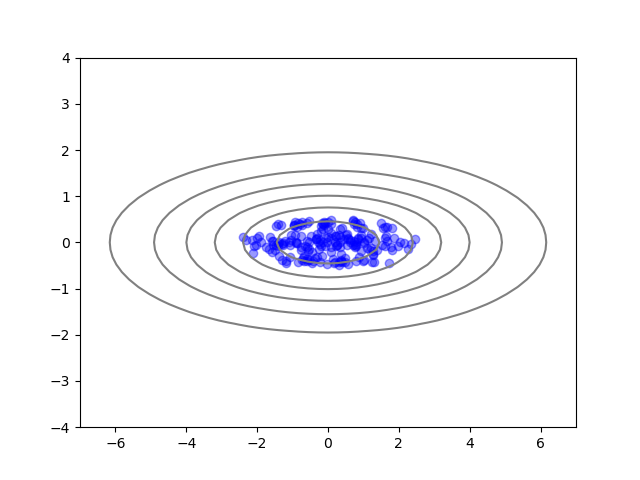}}}
    \subfloat[\centering BRWP $T=0.25$]{{\includegraphics[height=2.5cm]{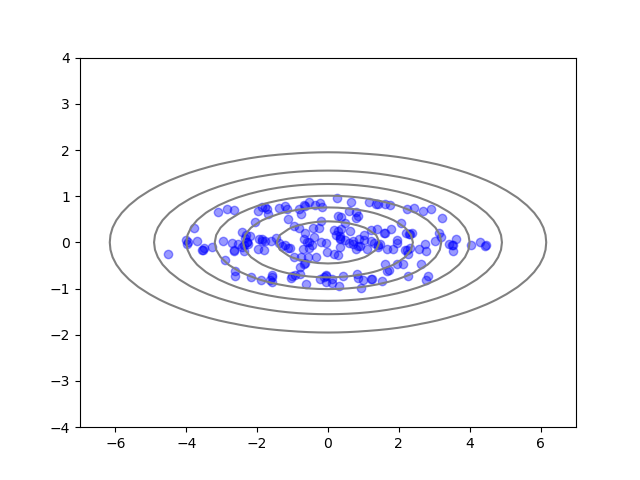}}}
    \subfloat[\centering BRWP $T=0.5$]{{\includegraphics[height=2.5cm]{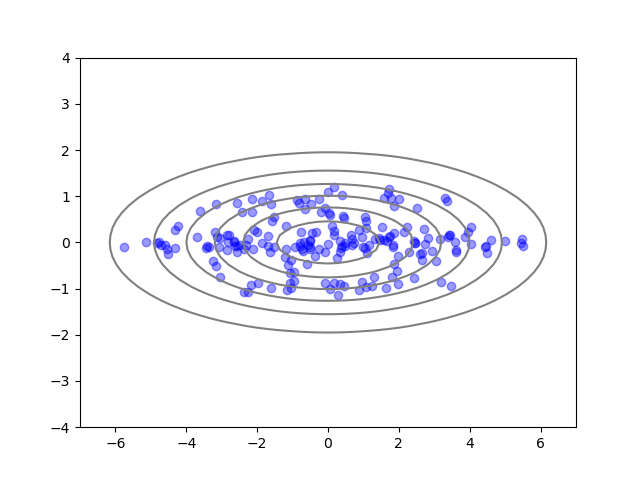}}}
    \subfloat[\centering BRWP $T=0.999$]{{\includegraphics[height=2.5cm]{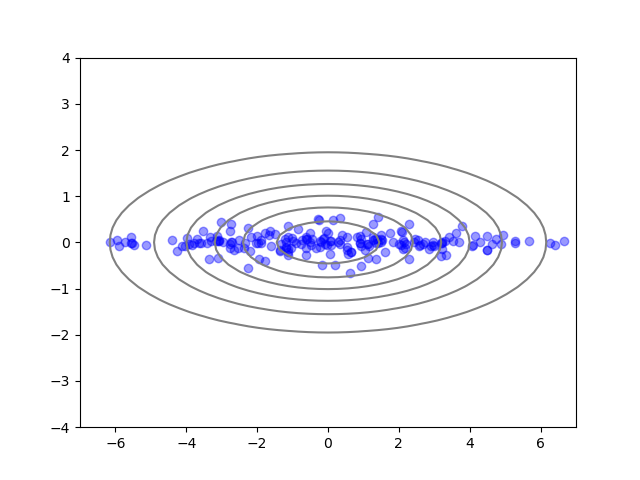}}}
    \caption{Particles after 200 iterations of BRWP with $T=0.05, 0.25, 0.5, 0.999$, applied to a two-dimensional Gaussian with condition number $\kappa=10$. Figures (a-d) in the top row have $N=1000$ samples, while figures (e-h) in the bottom row have $N=200$ samples. We observe that in higher dimensions, having fewer samples results in partial mode collapse as $T\rightarrow 0$. Moreover, for $T$ close to 1, the variance in the vertical direction of $\lambda(\Sigma)=1$ is reduced, demonstrating the bias of BRWP as described in \Cref{sec:Sec2}. }
    \label{fig:gaussian_5}
\end{figure}

\textbf{Variance reduction/mode collapse phenomenon.} \Cref{fig:gaussian_5} considers a 5-dimensional Gaussian, projected onto the first and last dimensions with target covariance 10 and 1, respectively, using $N=1000$ and $N=200$ samples. In the case of sufficiently many samples $N=1000$, we observe the same structural phenomenon as in \Cref{fig:gaussian_2_1000}. However, in the case where $N=200$, we observe a sample clustering phenomenon. For small values of $T$, the samples cluster more strongly around the true minimizer of $V$, which is the origin. For larger values of $T$, we observe that this clustering phenomenon is weaker, but there is bias due to the approximation as suggested in \Cref{sec:Sec2}.

The variance reduction suggests that the error incurred by approximating the distribution after each forward iteration by the empirical measure plays an effect in the convergence behavior. In the case where $T$ is small, this can be partially explained by the quadratic term dominating $V$ in the score formulation. 

\subsection{Gaussian Mixture}
To further illustrate the structure phenomenon, we can also use a mixture of Gaussians. Using the experiment setup in \citep{dwivedi2018log}, we consider sampling from the target density, given by a mixture of Gaussians $\gN(a, I)$ and $\gN(-a, I)$:
\begin{equation*}
    p(x) = \frac{1}{2(2\pi)^{d/2}} \left(e^{-\|x-a\|_2^2/2}+e^{-\|x+a\|_2^2/2}\right).
\end{equation*}
The corresponding potential is given by
\begin{subequations}
    \begin{gather}
        V(x) = \frac{1}{2}\|x-a\|_2^2 - \log\left(1+e^{-2x^\top a}\right), \\
        \nabla V(x) = x-a+2a(1+e^{2x^\top a})^{-1}.
    \end{gather}
\end{subequations}
We consider the same problem parameters as in \citet{dwivedi2018log,dalalyan2017theoretical}, taking dimension $d=2$ and the parameter $a = (1/2, 1/2)$. This gives strong convexity parameter $m=\frac{1}{2}$ and Lipschitz constant $L=1$. The initial distribution is chosen as $\gN(0, L^{-1}I) = \gN(0, I)$, and we initialize 200 particles with this distribution. For consistency, we use the same initialization for each of the compared methods. We compare with BRWP with parameters $T=0.01, 0.1$, with $P=25$ Monte Carlo samples for approximating the normalizing constant $\gZ$.

\begin{figure}
    \centering
    \subfloat[\centering ULA (50)]{{\includegraphics[height=2.5cm]{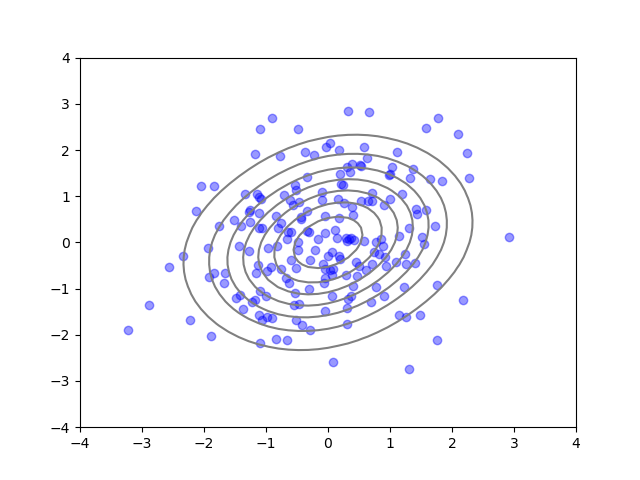}}}
    \subfloat[\centering ULA (500)]{{\includegraphics[height=2.5cm]{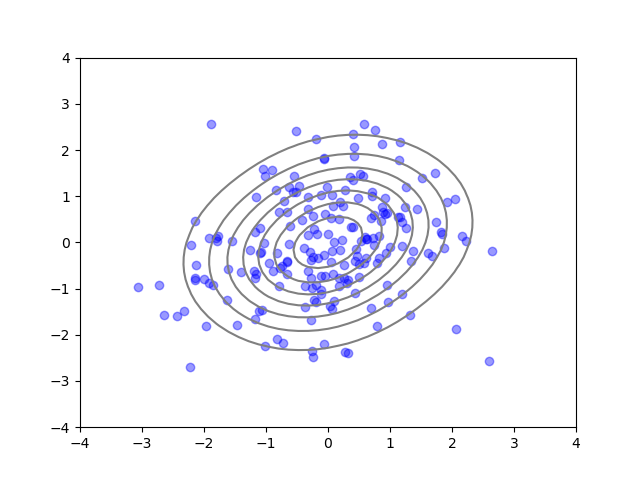} }}
    \subfloat[\centering MALA (50)]{{\includegraphics[height=2.5cm]{figs/mix_gaussian/ULA/50_zoom.png}}}
    \subfloat[\centering MALA (500)]{{\includegraphics[height=2.5cm]{figs/mix_gaussian/ULA/500_zoom.png}}}

    \subfloat[\centering BRWP\textsuperscript{1} (50)]{{\includegraphics[height=2.5cm]{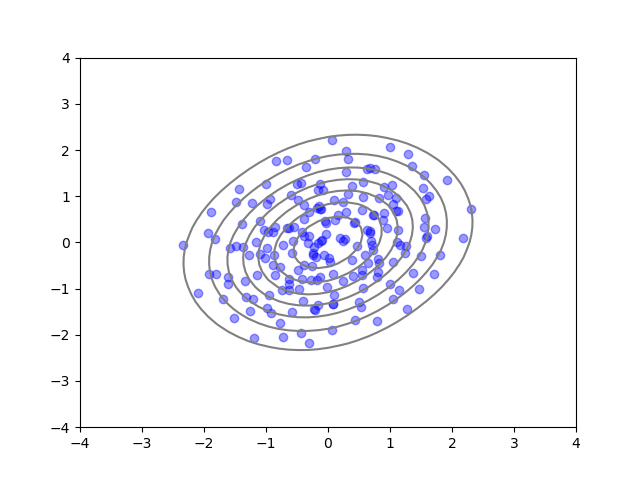}}}
    \subfloat[\centering BRWP\textsuperscript{1} (500)]{{\includegraphics[height=2.5cm]{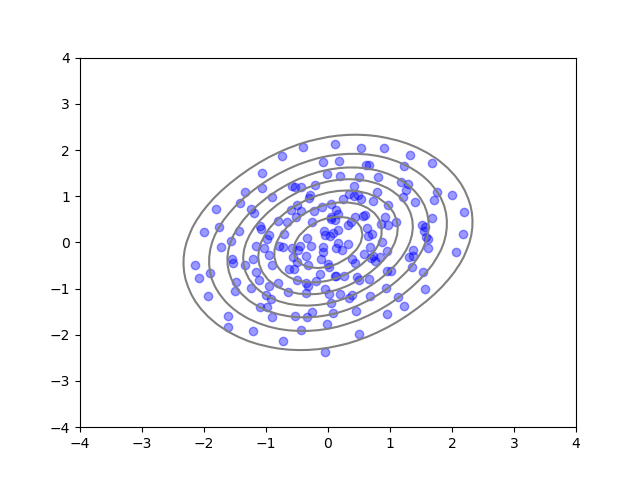} }}
    \subfloat[\centering BRWP\textsuperscript{2} (50)]{{\includegraphics[height=2.5cm]{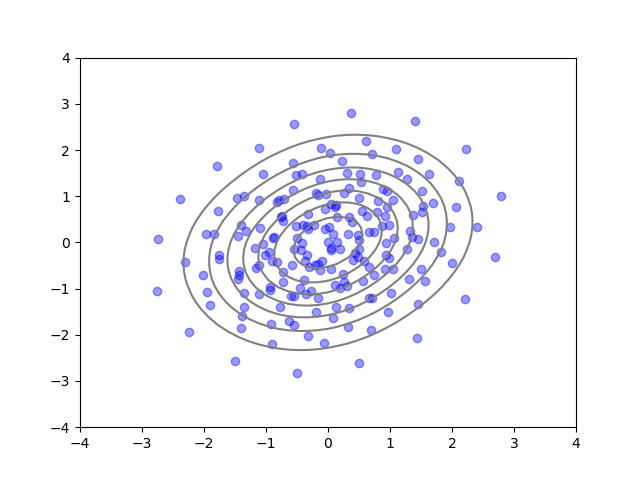}}}
    \subfloat[\centering BRWP\textsuperscript{2} (500)]{{\includegraphics[height=2.5cm]{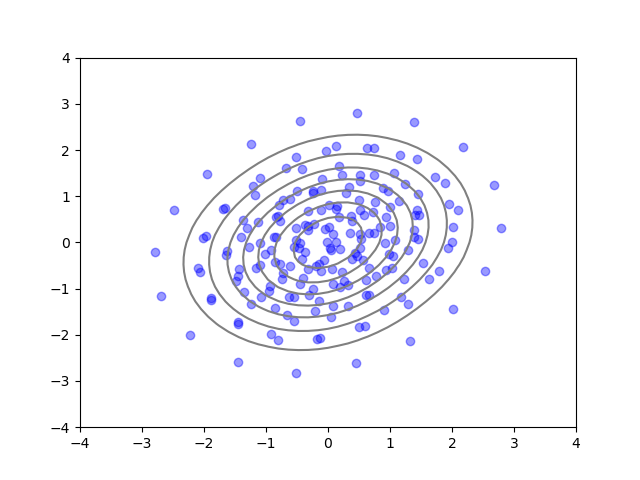}}}
    
    \caption{Evolution of particles under ULA, MALA, and BRWP for the bimodal distribution, with step-size $\eta=0.1$. Superscripts indicate different parameters of $T$, with BRWP\textsuperscript{1} having $T=0.01$ and BRWP\textsuperscript{2} having $T=0.1$. We observe that the iterates of BRWP converge in an organized manner as opposed to the randomness of ULA and MALA, with the lower level of $T$ giving some variance reduction properties.}
    \label{fig:mix_gaussian_evolution}
\end{figure}

We observe in \Cref{fig:mix_gaussian_evolution} that the samples of BRWP for parameters $T=0.01$ and $T=0.1$ both converge to roughly ellipsoidal patterns for this non-Gaussian case, fitting the level sets of the density. Moreover, we observe that the samples themselves exhibit some sort of structure, and do not have random-walk-like movements between the iterations as a result of the deterministic discretization.

\subsection{Bimodal Distribution}
As a more complicated toy example, we consider the two-dimensional bi-modal distribution as in \cite{wang2022accelerated}. This objective function has the form
\begin{equation*}
    p(x) \propto \exp(-2(\|x\|-3)^2) \left[\exp(-2(x_1-3)^2) + \exp(-2(x_1+3)^2)\right].
\end{equation*}
This is generated by the potential $V$ with gradient $\nabla V$ as follows:
\begin{subequations}
\begin{align}
        V(x) &= 2 (\|x\|-3)^2 - 2\log \left[\exp(-2(x_1-3)^2) + \exp(-2(x_1+3)^2)\right],\\
        \nabla V(x) &= 4  \frac{(\|x\|-3)x}{\|x\|} + \frac{4(x_1-3) \exp(-2(x_1-3)^2) +4(x_1+3) \exp(-2(x_1+3)^2)}{\exp(-2(x_1-3)^2) + \exp(-2(x_1+3)^2)} e_1,
\end{align}
\end{subequations}
where $e_1 = (1,0)^\top$ is the first standard coordinate vector. 
We fix the step-size for ULA and MALA to be $\eta=0.01$, and regularization parameter $T=0.01, 0.05, 0.1$ for the BRWP method. The samples are initialized as standard Gaussian $\gN(0,I)$, and we use 200 particles for simulation.

\begin{figure}
    \centering
    \subfloat[\centering ULA (10)]{{\includegraphics[height=2.5cm]{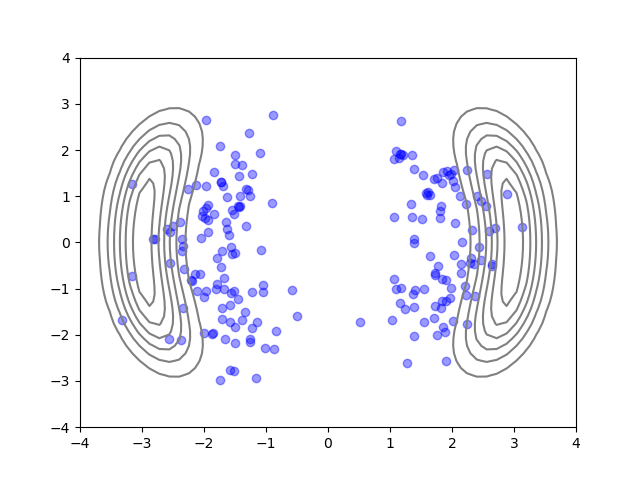}}}
    \subfloat[\centering ULA (50)]{{\includegraphics[height=2.5cm]{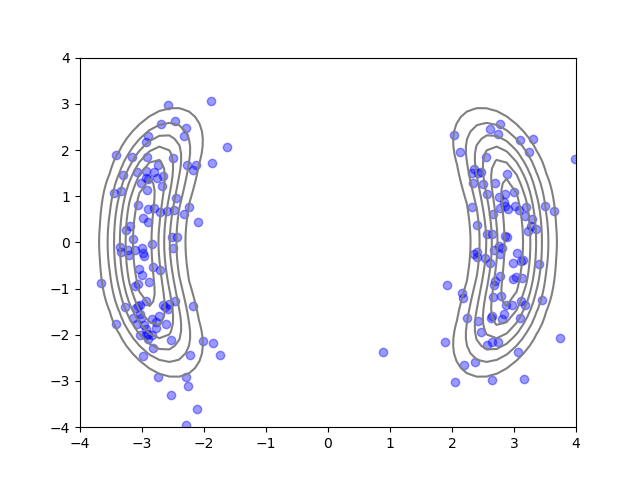}}}
    \subfloat[\centering ULA (100)]{{\includegraphics[height=2.5cm]{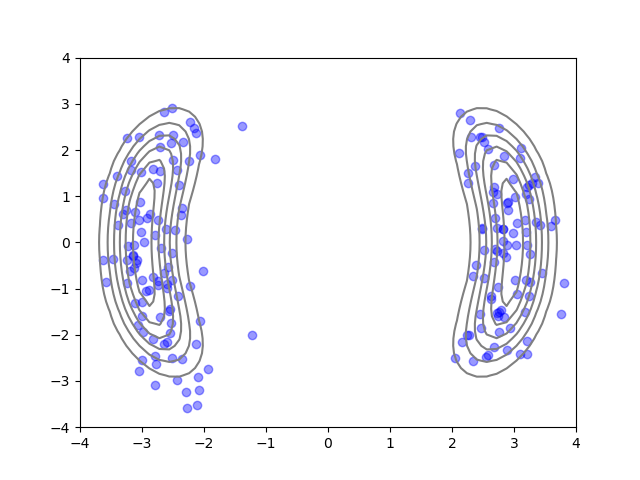}}}
    \subfloat[\centering ULA (2000)]{{\includegraphics[height=2.5cm]{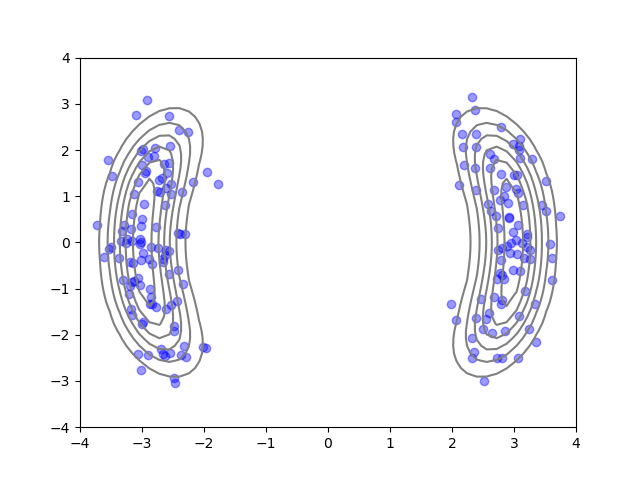}}}
    
    \subfloat[\centering MALA (10)]{{\includegraphics[height=2.5cm]{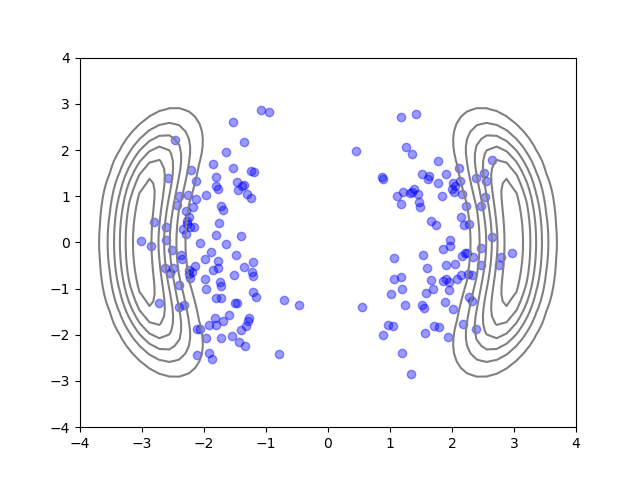}}}
    \subfloat[\centering MALA (50)]{{\includegraphics[height=2.5cm]{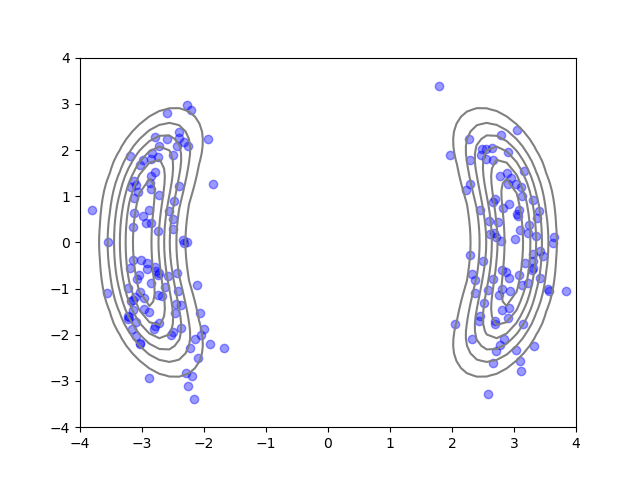}}}
    \subfloat[\centering MALA (100)]{{\includegraphics[height=2.5cm]{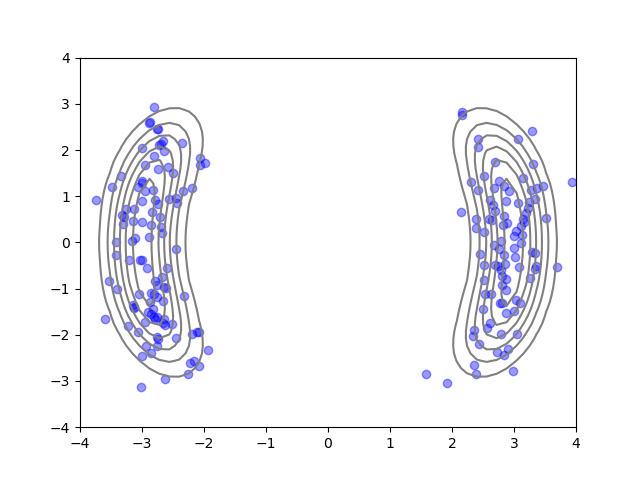}}}
    \subfloat[\centering MALA (2000)]{{\includegraphics[height=2.5cm]{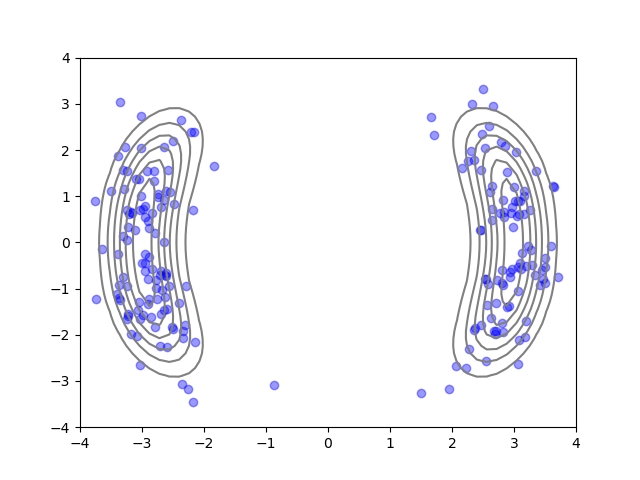}}}
    
    \subfloat[\centering BRWP (10)]{{\includegraphics[height=2.5cm]{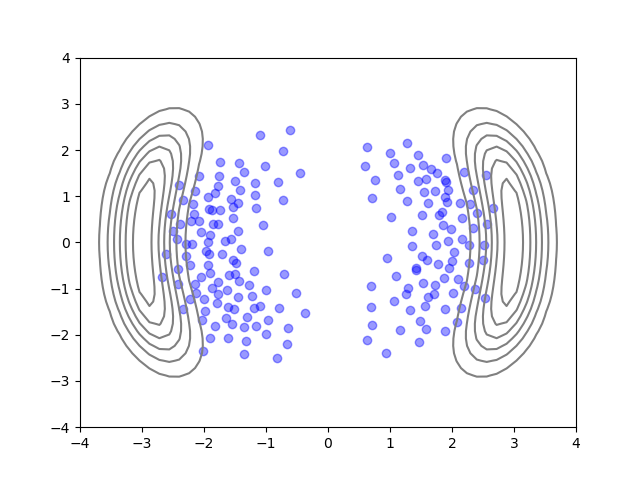}}}
    \subfloat[\centering BRWP (50)]{{\includegraphics[height=2.5cm]{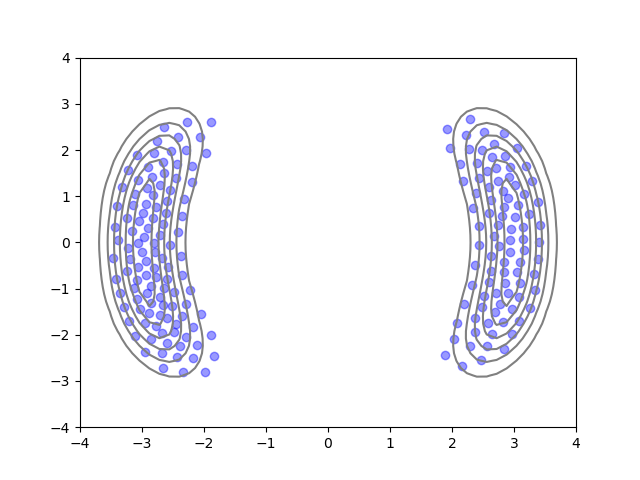}}}
    \subfloat[\centering BRWP (100)]{{\includegraphics[height=2.5cm]{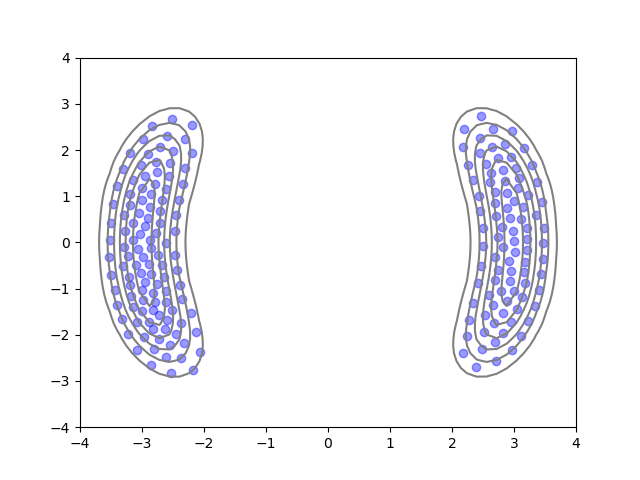}}}
    \subfloat[\centering BRWP (2000)]{{\includegraphics[height=2.5cm]{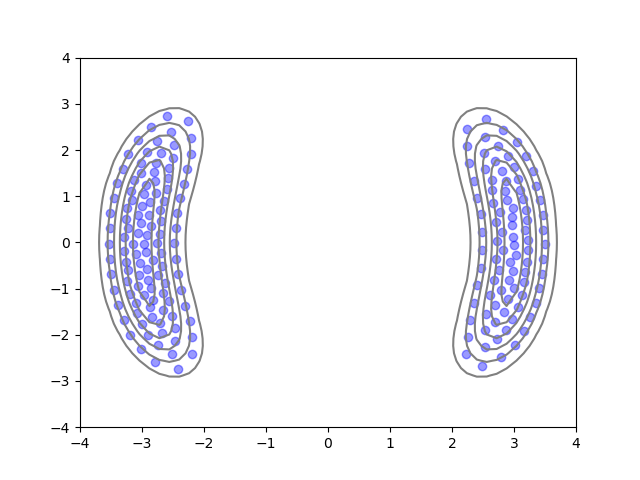}}}
    
    \caption{Evolution of particles under ULA, MALA and BRWP for the bimodal distribution, with step-size $\eta=0.01$. The parameter of $T$ was taken to be $T=0.01$ for BRWP. We observe that the iterates of BRWP converge in a structured manner to fit the distribution, and the iterates stay almost identical from 100 to 200 iterations. In contrast, the SDE based methods ULA and MALA have samples that continue to be random.}
    \label{fig:double_banana_evolution}
\end{figure}

In \Cref{fig:double_banana_evolution}, we plot the evolution of ULA, MALA and BRWP with $T=0.01$ at iteration numbers 10, 50, 100 and 2000. This figure illustrates that the samples of BRWP travel in a structured manner, and indeed stay approximately the same even after many iterations. In contrast, ULA and MALA continue to exhibit random behaviors after reaching the neighborhoods of the modes.

\begin{figure}
    \centering
    \subfloat[\centering ULA]{{\includegraphics[height=2.5cm]{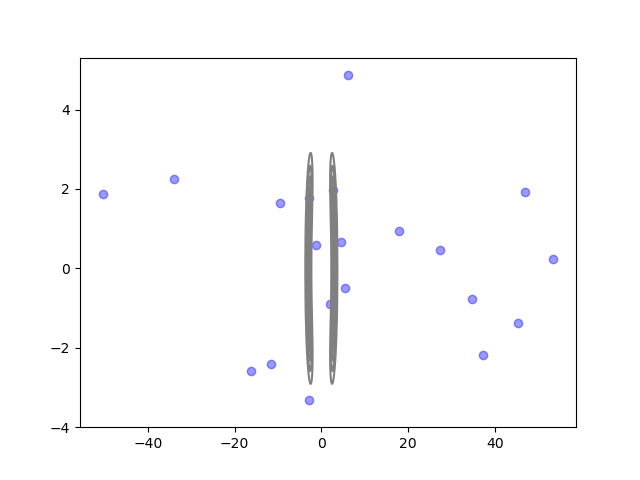}}}
    \subfloat[\centering MALA]{{\includegraphics[height=2.5cm]{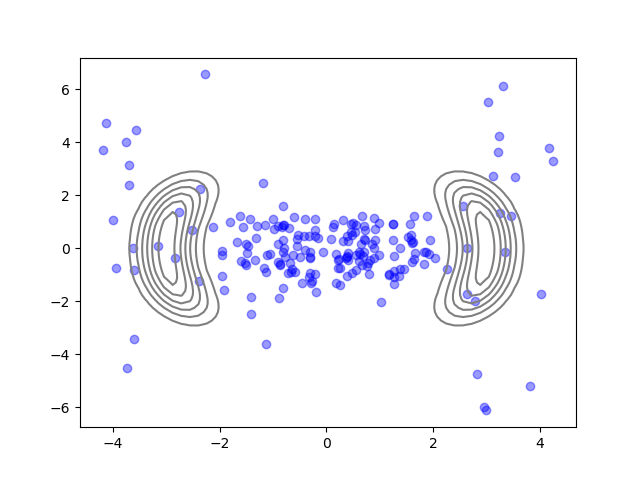}}}
    \subfloat[\centering BRWP $T=0.1$]{{\includegraphics[height=2.5cm]{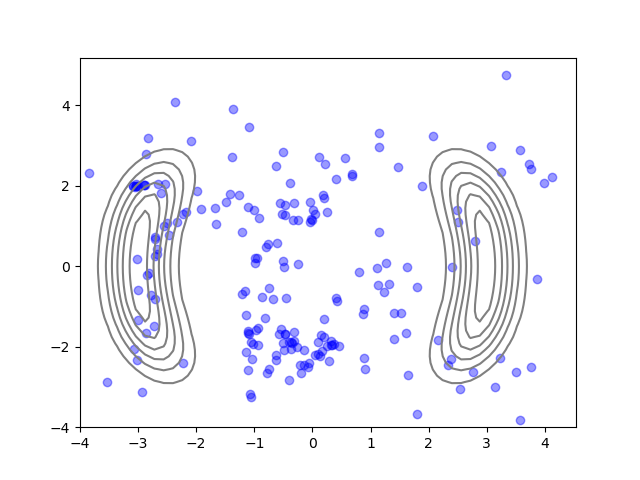}}}
    \subfloat[\centering BRWP $T=0.2$]{{\includegraphics[height=2.5cm]{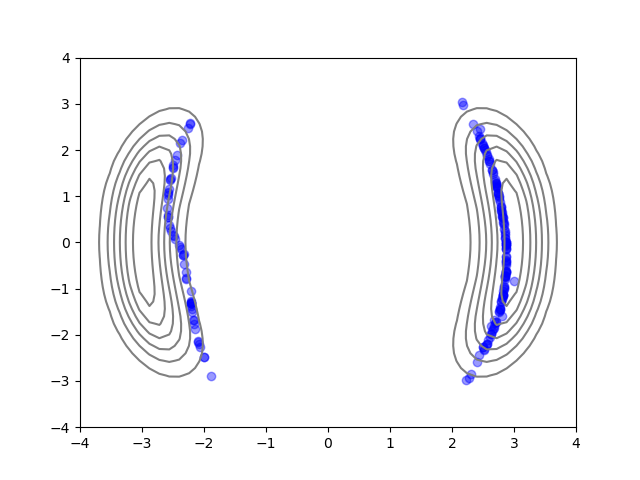}}}

    \subfloat[\centering MALA]{{\includegraphics[height=2.5cm]{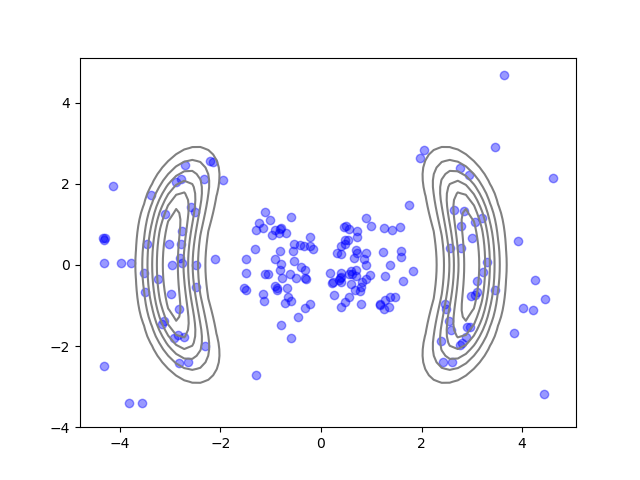}}}
    \subfloat[\centering BRWP $T=0.1$]{{\includegraphics[height=2.5cm]{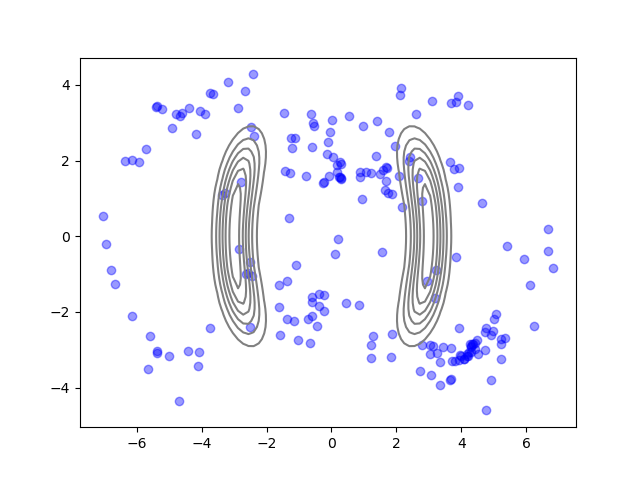}}}
    \subfloat[\centering BRWP $T=0.2$]{{\includegraphics[height=2.5cm]{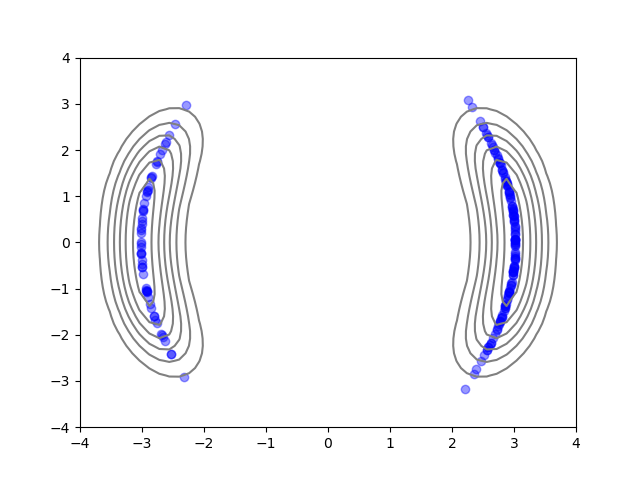}}}
    \caption{Evolution of particles under ULA, MALA and BRWP for the bimodal distribution, with very large step-size $\eta=0.5$. Figures (a-d) are at iteration 10, while (e-g) are at iteration 100. ULA diverges after 10 iterations. MALA and BRWP with $T=0.1$ do not converge to the modes. For sufficiently large $T$, BRWP removes low covariance components and allows for a larger step-size.}
    \label{fig:double_banana_bigSS}
\end{figure}

\Cref{fig:double_banana_bigSS} explores the behavior of the compared algorithms in the very large step-size regime, where none of the methods are expected to converge. Taking the step-size $\eta=0.5$, we have that ULA diverges, while MALA and BRWP with $T=0.1$ do not converge to neighborhoods of the modes. However, once we take $T=0.2$ to be sufficiently large, we again observe a convergent behavior. The iterates converge towards a curve that follows the valleys of $V$. This suggests that $T$ implicitly performs a variance reduction even in the case where the target distribution is not log-concave. 


\subsection{Bayesian Logistic Regression}
We additionally explore the performance for Bayesian logistic regression, in the framework detailed in \citet{dwivedi2018log,dalalyan2017theoretical}. The problem is as follows. Suppose that we have covariates $x \in \R^d$ as well as a binary variable $y \in \{0,1\}$. The logistic model is for the conditional distribution of $y$ given $x$ for a parameter $\theta \in \R^d$ is 
\begin{equation*}
    \sP(y = 1 \mid x, \theta) = \frac{\exp(\theta^\top x)}{1+\exp(\theta^\top x)}.
\end{equation*}

Given a binary vector $Y \in \{0,1\}^n$ and a feature matrix $X \in \R^{n \times d}$ with rows $x_i \in \R^d$, suppose we impose a prior density $\theta \sim \gN(0, \Sigma_X)$, where $\Sigma_X = \frac{1}{n}X^\top X$ is the sample covariance matrix of $X$. Then the posterior density of $\theta$ is given by
\begin{equation}
    p(\theta \mid X, Y) \propto \exp \left\{Y^\top X \theta - \sum_{i=1}^n \log\left(1+\exp(\theta^\top x_i)\right) - \alpha \|\Sigma_X^\frac{1}{2}\theta\|_2^2\right\},
\end{equation}
where $\alpha>0$ is a regularization parameter. This can be cast into the problem of sampling from a Gibbs distribution, with potential
\begin{align*}
    V(\theta) &= -Y^\top X \theta + \sum_{i=1}^{n} \log \left(1+\exp(\theta^\top x_i)\right) + \alpha \|\Sigma_X^{\frac{1}{2}} \theta\|_2^2, \\
    \nabla V(\theta) &= -X^\top Y + \sum_{i=1}^n \frac{x_i}{1+\exp(-\theta^\top x_i)} + \alpha \Sigma_X \theta.
\end{align*}
As in \citet{dwivedi2018log}, the eigenvalues of the Hessian are bounded by $L = (0.25 n + \alpha) \lambda_{\max}(\Sigma_X)$ and $m=\alpha \lambda_{\min}(\Sigma_X)$. In our experiments, we choose the logistic regression parameters as $\alpha =0.5, d=2, n=50$. We fix the step-size to be $\eta = 0.05$, and run each of the methods for 5000 iterations. We initialize $N=1000$ samples using the distribution $\gN(0, L^{-1} I)$ as in \citet{dwivedi2018log}.

For evaluation, we consider the error with respect to the true minimizers of $V$, denoted by $\theta^*$. This is also known as the maximum a posteriori (MAP) estimate in the Bayesian optimization literature. To compute $\theta^*$, we run gradient descent for 1000 iterations with step-size 1e-3, followed by 1000 iterations with step-size 1e-4, initialized at $\theta = (1,1)^\top$. For the computed samples, we compute the expected $\ell_1$ deviation from $\theta^*$ divided by $d$, as well as the $\ell_1$ distance of the sample mean to $\theta^*$ divided by $d$. The metrics are, where $\hat\theta_k$ is the empirical distribution and $\bar\theta$ is the sample mean at the $k$-th iteration,
\begin{equation}\label{eq:BLRMetrics}
    \varepsilon_1 = \frac{1}{d}\|\bar\theta - \theta^*\|_1,\, \varepsilon_2 = \frac{1}{d}\mathbb{E}\|\hat\theta_k - \theta^*\|_1.
\end{equation}
These metrics deviate from that of \citet{dwivedi2018log} in the sense that $\theta^*$ is chosen to be the minimum of $V$, instead of the $\theta=(1,1)$ used to generate the samples. This compensates for the bias generated by the added regularization, and makes it easier to compare the posterior means to the MAP estimate.

\begin{figure}
    \centering
    \subfloat[\centering $\varepsilon_1$]{{\includegraphics[height=5cm]{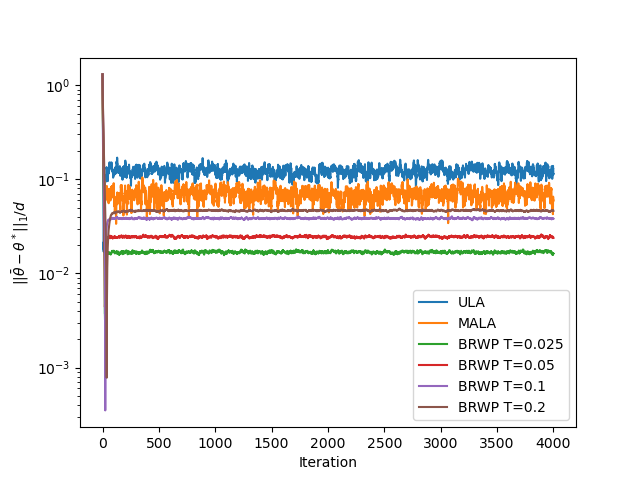}}}
    \subfloat[\centering $\varepsilon_2$]{{\includegraphics[height=5cm]{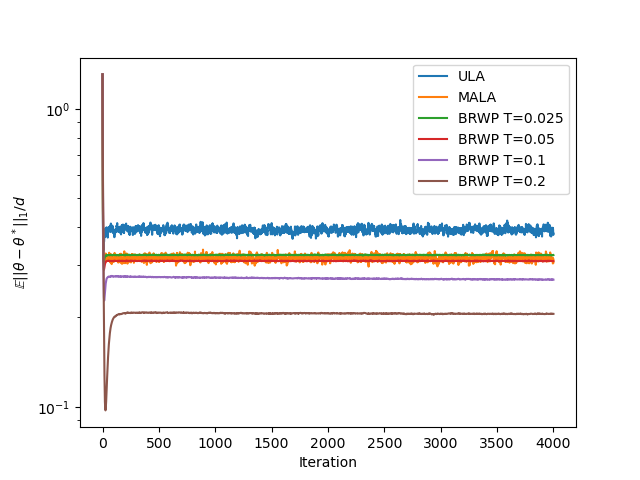}}}
    \caption{Plot of the $\varepsilon_1$ and $\varepsilon_2$ metrics. The regularization parameter is $\alpha=0.5$, with condition number $\kappa\approx28.2$. The step-size is $\eta=0.05$ for all methods. We observe that for small values of $T$, the sample mean is closer to the true parameter value $\theta^*$; for larger $T$, the variance is lower. This demonstrates a bias-variance trade-off of $T$. }
    \label{fig:bayesianLRErr}
\end{figure}

\begin{figure}
    \centering
    \subfloat[\centering ULA]{{\includegraphics[height=2.5cm,trim={0 0 0 0.7cm},clip]{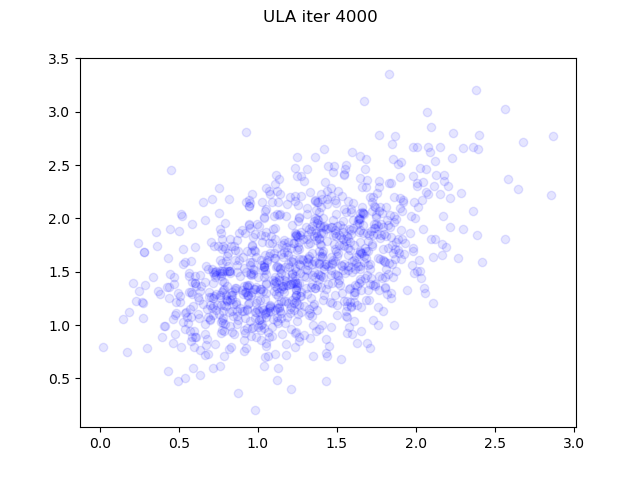}}}
    \subfloat[\centering MALA]{{\includegraphics[height=2.5cm,trim={0 0 0 0.7cm},clip]{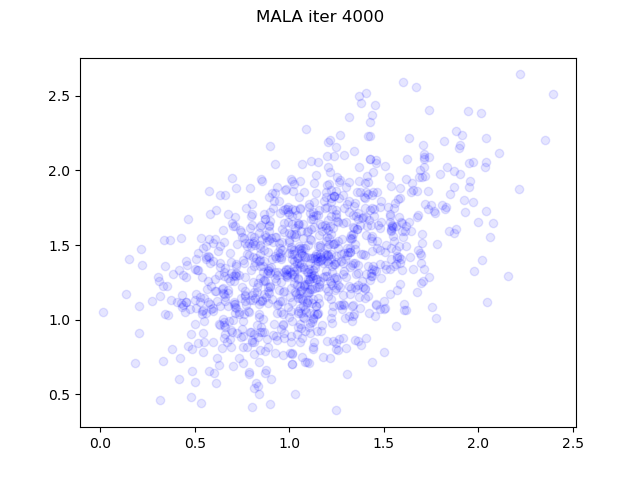}}}
    \subfloat[\centering BRWP $T=0.025$]{{\includegraphics[height=2.5cm,trim={0 0 0 0.7cm},clip]{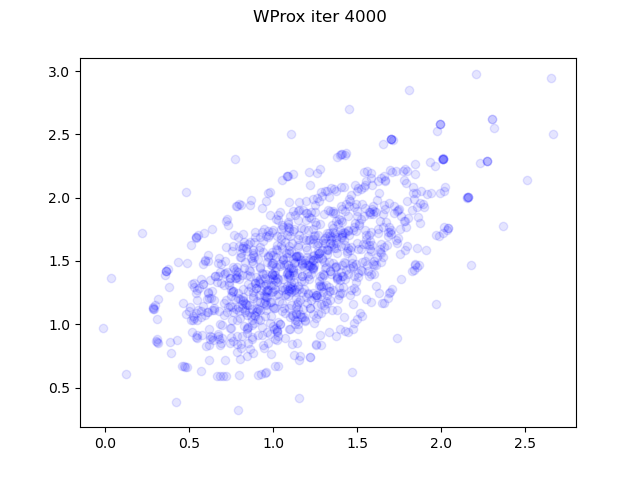}}}
    \subfloat[\centering BRWP $T=0.1$]{{\includegraphics[height=2.5cm,trim={0 0 0 0.7cm},clip]{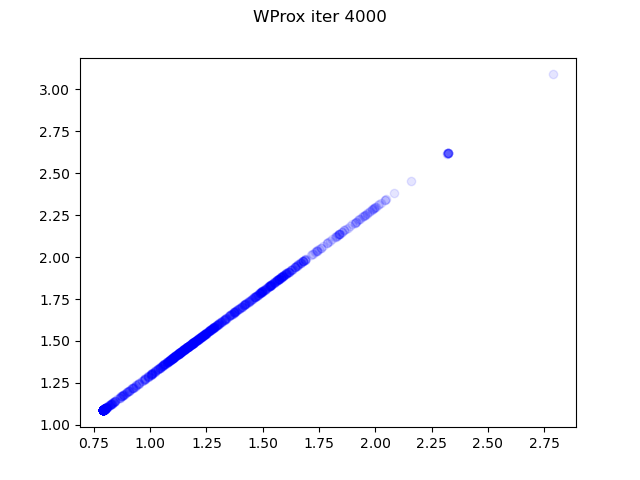}}}
    
    \caption{Plots of the samples of $\theta$ after 4000 iterations, with $N=1000$ samples. Parameters are $\alpha=0.5, \eta=0.05$. For this particular instantiation, we find that $\theta^* \approx (1.16, 1.45)$. We observe that for small $T$, we have a teardrop shaped structure. For large $T$, we have mode collapse in one direction.}
    \label{fig:bayesianLRSamples0.5}
\end{figure}

\begin{figure}
    \centering
    \subfloat[\centering BRWP $T=0.025$]{{\includegraphics[height=2.5cm,trim={0 0 0 0.7cm},clip]{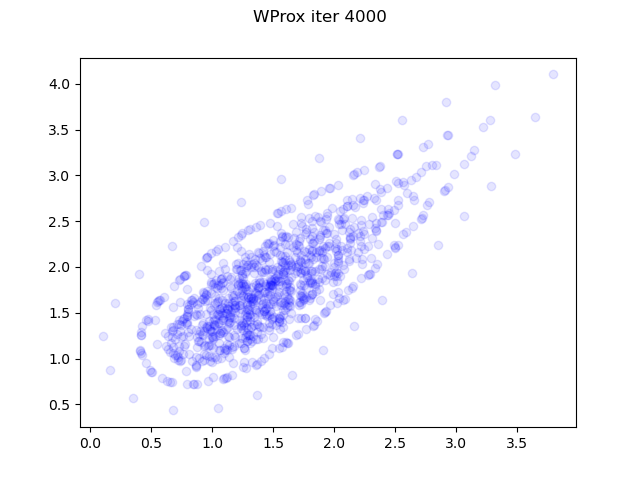}}}
    \subfloat[\centering BRWP $T=0.05$]{{\includegraphics[height=2.5cm,trim={0 0 0 0.7cm},clip]{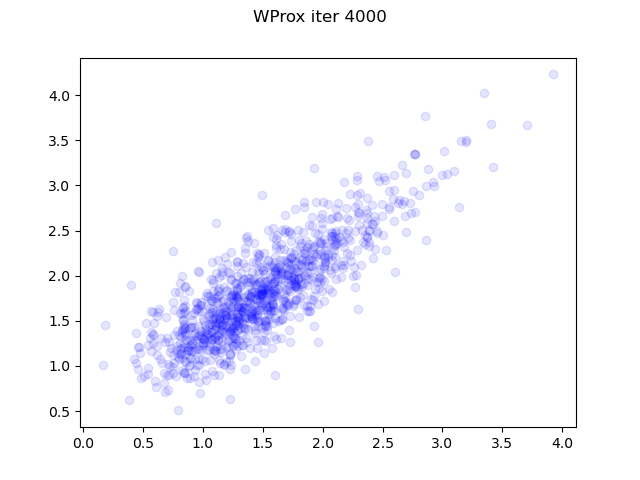}}}
    \subfloat[\centering BRWP $T=0.1$]{{\includegraphics[height=2.5cm,trim={0 0 0 0.7cm},clip]{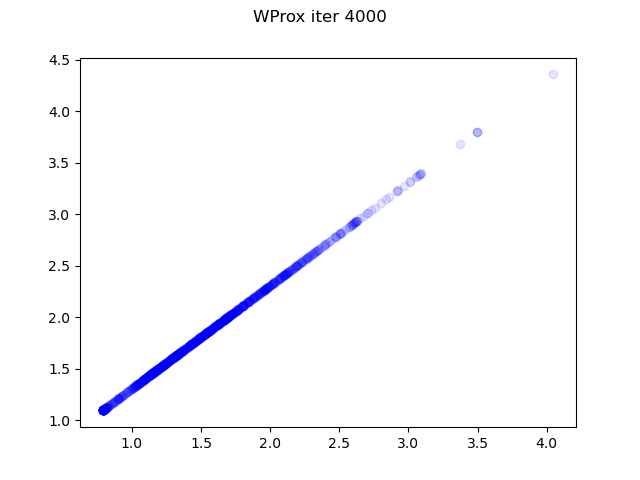}}}
    \subfloat[\centering BRWP $T=0.2$]{{\includegraphics[height=2.5cm,trim={0 0 0 0.7cm},clip]{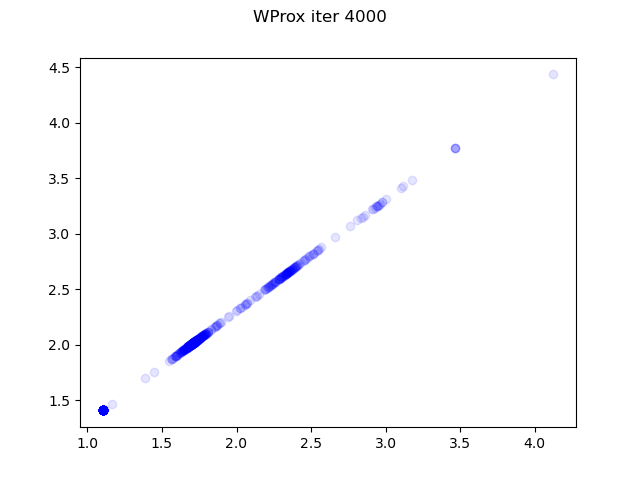}}}
    
    \caption{Plots of the samples of $\theta$ after 4000 iterations, with $N=1000$ samples. Parameters are $\alpha=0.1, \eta=0.05$. For this particular instantiation, we find that $\theta^* \approx (1.32, 1.62)$. We observe variance reduction in the approximately $y=-x$ direction as $T$ increases.}
    \label{fig:bayesianLRSamples0.1}
\end{figure}

\Cref{fig:bayesianLRErr} plots the error metrics $\varepsilon_1, \varepsilon_2$ as defined in \Cref{eq:BLRMetrics}. We observe that the metrics for the BRWP scheme for regularization parameters $T=0.025, 0.05, 0.1, 0.2$ are lower than ULA and MALA. Moreover, we observe that the error metrics converge after around 200 iterations and have significantly less noise across the iterations. From $\varepsilon_1$ being smaller, we have that the posterior mean is closer to the MAP estimate for BRWP. $\varepsilon_2$ being smaller demonstrates again the variance reduction of the scheme, with larger values of $T$ corresponding to less variance. \Cref{fig:bayesianLRSamples0.5,fig:bayesianLRSamples0.1} plot the samples after 4000 iterations for the various levels of $T$. We observe that for $T=0.025, 0.05$, a clear teardrop-shaped structure arises, traced out by the outer samples. For $T=0.1,0.2$, the samples appear collinear. 

We can additionally interpret $\varepsilon_2$ as an optimization objective, rather than a sampling objective. Using this interpretation, we have that for larger $T$, the optimization effect on $V$ is larger and dominates the diffusion. For sampling schemes such as ULA and MALA, this would be dictated by the regularization parameter $\beta$. For a convex objective $V$, as $\beta \rightarrow 0$, we have less diffusion effects, and the target density $\exp(-V(x)/\beta)$ converges to the Dirac mass at the minimizer of $V$.

\subsection{Bayesian Neural Network Training}
In this subsection, we consider the task of Bayesian neural network regression over the UCI dataset\footnote{\url{https://archive.ics.uci.edu/datasets}}. We consider the same setting given in \cite{wang2022accelerated,wang2019stein}. The task is to train neural networks to minimize the negative log-likelihood, where the likelihood is modelled as Gaussian. We consider the proposed BRWP method with the results reported in \cite{wang2022accelerated} with competing methods, namely accelerated information gradient flow (AIG), Wasserstein gradient flow (W-GF), and Stein variational gradient descent (SVGD) \citep{liu2016stein}. 

The neural networks are taken to be two-hidden-layer ReLU neural networks, with 50 activations in the hidden layers, initialized using the default Gaussian initialization. We take the same epoch and batch-size hyperparameters as \cite{wang2022accelerated}. For the step-size $\eta$ and Wasserstein regularization parameter $T$, we choose them by running a grid search over $\{1,2,5\} \times 10^{-i}$ for $i\in\{2,3,4,5\}$, to minimize RMSE. Each method was run using $N=10$ particles, each corresponding to a single neural network. The datasets are split with 90\% for training and 10\% for testing. We run 20 experiments for each dataset, with different dataset splits and initializations for each experiment. We report the root-mean-squared error (RMSE) and the test log-likelihood in \Cref{tab:RMSE,tab:BNN_llhood}. The values reported are the average over 20 independent experiments, with the population variance in subscripts. 

We observe that the numerical results are competitive with existing classical methods, with significantly higher variance. The high variance can be interpreted in multiple ways. Firstly, BRWP is able to find very good solutions with better RMSE and test log-likelihood than the competing methods, which do not find these good solutions in the 20 independent trials. This suggests better space exploration of BRWP around the high log-likelihood areas. The high variance may be caused by outliers caused by the kernel diffusion similar to \Cref{fig:bayesianLRSamples0.1}(c) and (d), which is cause for future work.

\begin{table}[]
\centering
\begin{tabular}{@{}crrrr@{}}
\toprule
Dataset & \multicolumn{1}{c}{BRWP} & \multicolumn{1}{c}{AIG} & \multicolumn{1}{c}{WGF} & \multicolumn{1}{c}{SVGD} \\ \midrule
Boston   & $3.309_{\pm 5.31\mathrm{e}-1}$ & $2.871_{\pm 3.41\mathrm{e}-3}$ & $3.077_{\pm 5.52\mathrm{e}-3}$ & $\pmb{2.775_{\pm 3.78\mathrm{e}-3}}$ \\
Combined & $\pmb{3.975_{\pm 3.94\mathrm{e}-2}}$ & $4.067_{\pm 9.27\mathrm{e}-1}$ & $4.077_{\pm 3.85\mathrm{e}-4}$ & ${4.070_{\pm 2.02\mathrm{e}-4}}$ \\
Concrete & $4.478_{\pm 2.05\mathrm{e}-1}$ & $\pmb{4.440_{\pm 1.34\mathrm{e}-1}}$ & $4.883_{\pm 1.93\mathrm{e}-1}$ & $4.888_{\pm 1.39\mathrm{e}-1}$ \\
Kin8nm   & $\pmb{0.089_{\pm 6.06\mathrm{e}-6}}$ & $0.094_{\pm 5.56\mathrm{e}-6}$ & $0.096_{\pm 3.36\mathrm{e}-5}$ & $0.095_{\pm 1.32\mathrm{e}-5}$ \\
Wine     & $0.623_{\pm 1.35\mathrm{e}-3}$ & $0.606_{\pm 1.40\mathrm{e}-5}$ & $0.614_{\pm 3.48\mathrm{e}-4}$ & $\pmb{0.604_{\pm 9.89\mathrm{e}-5}}$ \\ \bottomrule
\end{tabular}
\caption{Test root-mean-square-error (RMSE). Bold indicates smallest in row.}
\label{tab:RMSE}
\end{table}

\begin{table}[]
\begin{tabular}{@{}crrrr@{}}
\toprule
Dataset  & \multicolumn{1}{c}{BRWP} & \multicolumn{1}{c}{AIG}        & \multicolumn{1}{c}{WGF}        & \multicolumn{1}{c}{SVGD}       \\ \midrule
Boston   & $-2.629_{\pm 7.47\mathrm{e}-2}$                         & $\pmb{-2.609_{\pm1.34\mathrm{e}-4}}$ & $-2.694_{\pm2.83\mathrm{e}-4}$ & $-2.611_{\pm1.36\mathrm{e}-4}$  \\
Combined & $\pmb{-2.808_{\pm 3.45\mathrm{e}-3}}$                         & $-2.822_{\pm5.72\mathrm{e}-3}$ & $-2.825_{\pm2.36\mathrm{e}-5}$ & $-2.823_{\pm1.24\mathrm{e}-5}$ \\
Concrete & $-3.003_{\pm 2.51\mathrm{e}-2}$                         & $\pmb{-2.884_{\pm8.84\mathrm{e}-3}}$ & $-2.971_{\pm8.93\mathrm{e}-3}$ & $-2.978_{\pm6.05\mathrm{e}-3}$ \\
Kin8nm   & $\pmb{0.999_{\pm 9.53\mathrm{e}-4}}$                         & $0.951_{\pm6.43\mathrm{e}-4}$  & $0.923_{\pm3.37\mathrm{e}-3}$  & $0.932_{\pm1.43\mathrm{e}-3}$  \\
Wine     & $\pmb{-0.947_{\pm 4.21\mathrm{e}-3}}$                         & $-0.961_{\pm1.28\mathrm{e}-4}$ & $-0.961_{\pm3.17\mathrm{e}-4}$ & $-0.952_{\pm9.89\mathrm{e}-5}$ \\ \bottomrule
\end{tabular}
\caption{Test log-likelihood. Bold indicates largest in row.}
\label{tab:BNN_llhood}
\end{table}

\section{Conclusion}
This work presents a novel deterministic approach to sampling using the regularized Wasserstein proximal. By approximating the density as a regularized Wasserstein proximal of the empirical distribution, we obtain a particle-based ODE approximation to the Fokker-Planck equation at each time step. Discretizing this approximate ODE using a backwards Euler step gives a deterministic sampling algorithm. We fully characterize the convergence and give closed-form iterations in the case of an Ornstein-Uhlenbeck process with quadratic potential, corresponding to a Gaussian target distribution. Moreover, we observe numerically that the proposed BRWP scheme converges in a visually structured manner by foregoing stochasticity. 

While the empirical results demonstrate the practicality of our scheme as an alternative to non-deterministic sampling algorithms such as ULA and MALA in the case of low-dimensional non-log-concave distributions, there are two main limitations. Firstly, the variance reduction/mode collapse phenomenon increases the number of samples required, and thus the complexity of the BRWP method. Secondly, the analysis is currently limited to the case of Gaussians. To further cement this method as a suitable and provably convergent method for sampling, demonstrating the convergence rate for more general distributions such as log-concave distributions is required. We conjecture that the variance reducing behavior of $T$ can be shown to implicitly reduce or remove deviations in directions of small covariance, which may be useful to remove small noise in data. Various open questions corresponding to the proposed scheme follow.

\textbf{Convergence rates for log-concave density.} The preliminary analysis given is only for Gaussian densities, with empirical results suggesting that the method continues to work. We believe that the closed-form updates of the BRWP scheme can lead to an analytic solution for convergence rates for log-concave target densities.

\textbf{Discretization and approximation error.} We made four approximating steps at the start of \Cref{sec:Sec2} to construct the BRWP scheme, including approximating the Fokker-Planck equation, ODE discretization using the backwards Euler method, and replacing densities with empirical measures at each iteration. The impact and convergence rate of these approximations with respect to the number of samples or the regularization parameters could be an interesting direction.

\textbf{Sample scaling in dimension and variance reduction phenomenon.} We observed in \Cref{sec:ExperimentGaussian} that in higher dimensions, the number of samples plays a role in variance reduction, even if the analytic rates are dimension independent. Quantifying or mitigating this effect for either sampling or optimization would be beneficial for high-dimensional applications.

\textbf{Structure of the iterates.} We observed empirically that the iterates cluster in a visually cohesive manner, with external iterates approximately lying on level sets of the density. However, this is deeply connected with the discretization method, and could prove to be a difficult yet rewarding problem.

\section*{Acknowledgements}
H.Y. Tan acknowledges support from GSK.ai, the Masason Foundation, and the European Union Horizon 2020 research and innovation programme under the Marie Skodowska-Curie grant agreement No. 777826 NoMADS. S. Osher is supported in part by AFOSR MURI FP 9550-18-1-502 and ONR grants: N00014-20-1-2093 and N00014-20-1-2787. W. Li's work is supported by AFOSR MURI FP 9550-18-1-502, AFOSR YIP award No. FA9550-23-1-0087, NSF DMS-2245097, and NSF RTG: 2038080. 


\bibliography{refs}

\begin{thebibliography}{53}
\providecommand{\natexlab}[1]{#1}
\providecommand{\url}[1]{\texttt{#1}}
\expandafter\ifx\csname urlstyle\endcsname\relax
  \providecommand{\doi}[1]{doi: #1}\else
  \providecommand{\doi}{doi: \begingroup \urlstyle{rm}\Url}\fi

\bibitem[Andrieu et~al.(2003)Andrieu, De~Freitas, Doucet, and
  Jordan]{andrieu2003introduction}
Christophe Andrieu, Nando De~Freitas, Arnaud Doucet, and Michael~I Jordan.
\newblock An introduction to {M}{C}{M}{C} for machine learning.
\newblock \emph{Machine learning}, 50:\penalty0 5--43, 2003.

\bibitem[Batzolis et~al.(2021)Batzolis, Stanczuk, Sch{\"o}nlieb, and
  Etmann]{batzolis2021conditional}
Georgios Batzolis, Jan Stanczuk, Carola-Bibiane Sch{\"o}nlieb, and Christian
  Etmann.
\newblock Conditional image generation with score-based diffusion models.
\newblock \emph{arXiv preprint arXiv:2111.13606}, 2021.

\bibitem[Baumgartner(2011)]{baumgartner2011inequality}
Bernhard Baumgartner.
\newblock An inequality for the trace of matrix products, using absolute
  values.
\newblock \emph{arXiv preprint arXiv:1106.6189}, 2011.

\bibitem[B{\'e}lisle et~al.(1993)B{\'e}lisle, Romeijn, and
  Smith]{belisle1993hit}
Claude~JP B{\'e}lisle, H~Edwin Romeijn, and Robert~L Smith.
\newblock Hit-and-run algorithms for generating multivariate distributions.
\newblock \emph{Mathematics of Operations Research}, 18\penalty0 (2):\penalty0
  255--266, 1993.

\bibitem[Benamou \& Brenier(2000)Benamou and Brenier]{benamou2000computational}
Jean-David Benamou and Yann Brenier.
\newblock A computational fluid mechanics solution to the {M}onge-{K}antorovich
  mass transfer problem.
\newblock \emph{Numerische Mathematik}, 84\penalty0 (3):\penalty0 375--393,
  2000.

\bibitem[Betancourt(2017)]{betancourt2017conceptual}
Michael Betancourt.
\newblock A conceptual introduction to {H}amiltonian {M}onte {C}arlo.
\newblock \emph{arXiv preprint arXiv:1701.02434}, 2017.

\bibitem[Bond-Taylor et~al.(2021)Bond-Taylor, Leach, Long, and
  Willcocks]{bond2021deep}
Sam Bond-Taylor, Adam Leach, Yang Long, and Chris~G Willcocks.
\newblock Deep generative modelling: A comparative review of {VAE}s, {GAN}s,
  normalizing flows, energy-based and autoregressive models.
\newblock \emph{IEEE transactions on pattern analysis and machine
  intelligence}, 2021.

\bibitem[Botev et~al.(2010)Botev, Grotowski, and Kroese]{botev2010kernel}
Zdravko~I Botev, Joseph~F Grotowski, and Dirk~P Kroese.
\newblock Kernel density estimation via diffusion.
\newblock \emph{Annals of Statistics}, 38\penalty0 (5):\penalty0 2916--2957,
  2010.

\bibitem[Brooks et~al.(2011)Brooks, Gelman, Jones, and
  Meng]{brooks2011handbook}
Steve Brooks, Andrew Gelman, Galin Jones, and Xiao-Li Meng.
\newblock \emph{Handbook of {M}arkov chain {M}onte {C}arlo}.
\newblock CRC press, 2011.

\bibitem[Carrillo et~al.(2019)Carrillo, Craig, and
  Patacchini]{carrillo2019blob}
Jos{\'e}~Antonio Carrillo, Katy Craig, and Francesco~S Patacchini.
\newblock A blob method for diffusion.
\newblock \emph{Calculus of Variations and Partial Differential Equations},
  58:\penalty0 1--53, 2019.

\bibitem[Chen et~al.(2018)Chen, Rubanova, Bettencourt, and
  Duvenaud]{chen2018neural}
Ricky~TQ Chen, Yulia Rubanova, Jesse Bettencourt, and David~K Duvenaud.
\newblock Neural ordinary differential equations.
\newblock \emph{Advances in neural information processing systems}, 31, 2018.

\bibitem[Dai et~al.(2021)Dai, Jiao, Kang, Lu, and Zhijian~Yang]{dai2021global}
Yin Dai, Yuling Jiao, Lican Kang, Xiliang Lu, and Jerry Zhijian~Yang.
\newblock Global optimization via {S}chr\"odinger-{F}\"ollmer diffusion.
\newblock \emph{arXiv e-prints}, pp.\  arXiv--2111, 2021.

\bibitem[Dalalyan(2017)]{dalalyan2017theoretical}
Arnak~S Dalalyan.
\newblock Theoretical guarantees for approximate sampling from smooth and
  log-concave densities.
\newblock \emph{Journal of the Royal Statistical Society Series B: Statistical
  Methodology}, 79\penalty0 (3):\penalty0 651--676, 2017.

\bibitem[Del~Moral(2013)]{del2013mean}
Pierre Del~Moral.
\newblock \emph{Mean field simulation for Monte Carlo integration}.
\newblock CRC press, 2013.

\bibitem[Devroye et~al.(2018)Devroye, Mehrabian, and Reddad]{devroye2018total}
Luc Devroye, Abbas Mehrabian, and Tommy Reddad.
\newblock The total variation distance between high-dimensional {G}aussians
  with the same mean.
\newblock \emph{arXiv preprint arXiv:1810.08693}, 2018.

\bibitem[Durmus \& Moulines(2019)Durmus and Moulines]{durmus2019high}
Alain Durmus and Eric Moulines.
\newblock High-dimensional {B}ayesian inference via the unadjusted {L}angevin
  algorithm.
\newblock 2019.

\bibitem[Dwivedi et~al.(2018)Dwivedi, Chen, Wainwright, and Yu]{dwivedi2018log}
Raaz Dwivedi, Yuansi Chen, Martin~J Wainwright, and Bin Yu.
\newblock Log-concave sampling: {M}etropolis-{H}astings algorithms are fast!
\newblock In \emph{Conference on learning theory}, pp.\  793--797. PMLR, 2018.

\bibitem[Gardiner et~al.(1985)]{gardiner1985handbook}
Crispin~W Gardiner et~al.
\newblock \emph{Handbook of stochastic methods}, volume~3.
\newblock Springer Berlin, 1985.

\bibitem[Gramacki(2018)]{gramacki2018nonparametric}
Artur Gramacki.
\newblock \emph{Nonparametric kernel density estimation and its computational
  aspects}, volume~37.
\newblock Springer, 2018.

\bibitem[Horn \& Johnson(2012)Horn and Johnson]{horn2012matrix}
Roger~A Horn and Charles~R Johnson.
\newblock \emph{Matrix analysis}.
\newblock Cambridge University Press, 2012.

\bibitem[Hyv{\"a}rinen \& Dayan(2005)Hyv{\"a}rinen and
  Dayan]{hyvarinen2005estimation}
Aapo Hyv{\"a}rinen and Peter Dayan.
\newblock Estimation of non-normalized statistical models by score matching.
\newblock \emph{Journal of Machine Learning Research}, 6\penalty0 (4), 2005.

\bibitem[Jordan et~al.(1998)Jordan, Kinderlehrer, and
  Otto]{jordan1998variational}
Richard Jordan, David Kinderlehrer, and Felix Otto.
\newblock The variational formulation of the {F}okker--{P}lanck equation.
\newblock \emph{SIAM journal on mathematical analysis}, 29\penalty0
  (1):\penalty0 1--17, 1998.

\bibitem[Karatzas \& Shreve(1991)Karatzas and Shreve]{karatzas1991brownian}
Ioannis Karatzas and Steven Shreve.
\newblock \emph{Brownian motion and stochastic calculus}, volume 113.
\newblock Springer Science \& Business Media, 1991.

\bibitem[Kardar(2007)]{kardar2007statistical}
Mehran Kardar.
\newblock \emph{Statistical physics of particles}.
\newblock Cambridge University Press, 2007.

\bibitem[Kim \& Scott(2012)Kim and Scott]{kim2012robust}
JooSeuk Kim and Clayton~D Scott.
\newblock Robust kernel density estimation.
\newblock \emph{The Journal of Machine Learning Research}, 13\penalty0
  (1):\penalty0 2529--2565, 2012.

\bibitem[Kubo(1963)]{kubo1963stochastic}
Ryogo Kubo.
\newblock Stochastic liouville equations.
\newblock \emph{Journal of Mathematical Physics}, 4\penalty0 (2):\penalty0
  174--183, 1963.

\bibitem[Laumont et~al.(2022)Laumont, Bortoli, Almansa, Delon, Durmus, and
  Pereyra]{laumont2022bayesian}
R{\'e}mi Laumont, Valentin~De Bortoli, Andr{\'e}s Almansa, Julie Delon, Alain
  Durmus, and Marcelo Pereyra.
\newblock {B}ayesian imaging using plug \& play priors: when langevin meets
  tweedie.
\newblock \emph{SIAM Journal on Imaging Sciences}, 15\penalty0 (2):\penalty0
  701--737, 2022.

\bibitem[Li et~al.(2023{\natexlab{a}})Li, Liu, Chen, Wu, Flynn, Ding, and
  Chen]{li2023reducing}
Wei Li, Wei Liu, Jinlin Chen, Libing Wu, Patrick~D Flynn, Wei Ding, and Ping
  Chen.
\newblock Reducing mode collapse with {M}onge--{K}antorovich optimal transport
  for generative adversarial networks.
\newblock \emph{IEEE Transactions on Cybernetics}, 2023{\natexlab{a}}.

\bibitem[Li et~al.(2023{\natexlab{b}})Li, Liu, and Osher]{li2023kernel}
Wuchen Li, Siting Liu, and Stanley Osher.
\newblock A kernel formula for regularized {W}asserstein proximal operators.
\newblock \emph{arXiv preprint arXiv:2301.10301}, 2023{\natexlab{b}}.

\bibitem[Liouville(1838)]{liouville1838note}
Joseph Liouville.
\newblock Note sur la th{\'e}orie de la variation des constantes arbitraires.
\newblock \emph{Journal de math{\'e}matiques pures et appliqu{\'e}es},
  3:\penalty0 342--349, 1838.

\bibitem[Liu \& Wang(2016)Liu and Wang]{liu2016stein}
Qiang Liu and Dilin Wang.
\newblock Stein variational gradient descent: A general purpose {B}ayesian
  inference algorithm.
\newblock \emph{Advances in neural information processing systems}, 29, 2016.

\bibitem[MacKay(1995)]{mackay1995bayesian}
David~JC MacKay.
\newblock {B}ayesian neural networks and density networks.
\newblock \emph{Nuclear Instruments and Methods in Physics Research Section A:
  Accelerators, Spectrometers, Detectors and Associated Equipment},
  354\penalty0 (1):\penalty0 73--80, 1995.

\bibitem[Maoutsa et~al.(2020)Maoutsa, Reich, and Opper]{maoutsa2020interacting}
Dimitra Maoutsa, Sebastian Reich, and Manfred Opper.
\newblock Interacting particle solutions of {F}okker--{P}lanck equations
  through gradient--log--density estimation.
\newblock \emph{Entropy}, 22\penalty0 (8):\penalty0 802, 2020.

\bibitem[Mattingly et~al.(2002)Mattingly, Stuart, and
  Higham]{mattingly2002ergodicity}
Jonathan~C Mattingly, Andrew~M Stuart, and Desmond~J Higham.
\newblock Ergodicity for {SDE}s and approximations: locally {L}ipschitz vector
  fields and degenerate noise.
\newblock \emph{Stochastic processes and their applications}, 101\penalty0
  (2):\penalty0 185--232, 2002.

\bibitem[Mengersen \& Tweedie(1996)Mengersen and Tweedie]{mengersen1996rates}
Kerrie~L Mengersen and Richard~L Tweedie.
\newblock Rates of convergence of the {H}astings and {M}etropolis algorithms.
\newblock \emph{The annals of Statistics}, 24\penalty0 (1):\penalty0 101--121,
  1996.

\bibitem[Meyn \& Tweedie(1994)Meyn and Tweedie]{meyn1994computable}
Sean~P Meyn and Robert~L Tweedie.
\newblock Computable bounds for geometric convergence rates of {M}arkov chains.
\newblock \emph{The Annals of Applied Probability}, pp.\  981--1011, 1994.

\bibitem[Nijkamp et~al.(2022)Nijkamp, Gao, Sountsov, Vasudevan, Pang, Zhu, and
  Wu]{nijkamp2022mcmc}
Erik Nijkamp, Ruiqi Gao, Pavel Sountsov, Srinivas Vasudevan, Bo~Pang, Song-Chun
  Zhu, and Ying~Nian Wu.
\newblock {MCMC} should mix: learning energy-based model with neural transport
  latent space {MCMC}.
\newblock In \emph{International Conference on Learning Representations (ICLR
  2022).}, 2022.

\bibitem[Osher et~al.(2023)Osher, Heaton, and Wu~Fung]{osher2023hamilton}
Stanley Osher, Howard Heaton, and Samy Wu~Fung.
\newblock A {H}amilton--{J}acobi-based proximal operator.
\newblock \emph{Proceedings of the National Academy of Sciences}, 120\penalty0
  (14):\penalty0 e2220469120, 2023.

\bibitem[Otto(2001)]{otto2001geometry}
Felix Otto.
\newblock The geometry of dissipative evolution equations: the porous medium
  equation.
\newblock 2001.

\bibitem[Parisi(1981)]{parisi1981correlation}
Giorgio Parisi.
\newblock Correlation functions and computer simulations.
\newblock \emph{Nuclear Physics B}, 180\penalty0 (3):\penalty0 378--384, 1981.

\bibitem[Patterson \& Teh(2013)Patterson and Teh]{patterson2013stochastic}
Sam Patterson and Yee~Whye Teh.
\newblock Stochastic gradient {R}iemannian {L}angevin dynamics on the
  probability simplex.
\newblock \emph{Advances in neural information processing systems}, 26, 2013.

\bibitem[Pereyra(2016)]{pereyra2016proximal}
Marcelo Pereyra.
\newblock Proximal {M}arkov chain {M}onte {C}arlo algorithms.
\newblock \emph{Statistics and Computing}, 26:\penalty0 745--760, 2016.

\bibitem[Roberts \& Tweedie(1996)Roberts and Tweedie]{roberts1996exponential}
Gareth~O Roberts and Richard~L Tweedie.
\newblock Exponential convergence of {L}angevin distributions and their
  discrete approximations.
\newblock \emph{Bernoulli}, pp.\  341--363, 1996.

\bibitem[Rossky et~al.(1978)Rossky, Doll, and Friedman]{rossky1978brownian}
Peter~J Rossky, Jimmie~D Doll, and Harold~L Friedman.
\newblock Brownian dynamics as smart {M}onte {C}arlo simulation.
\newblock \emph{The Journal of Chemical Physics}, 69\penalty0 (10):\penalty0
  4628--4633, 1978.

\bibitem[Song et~al.(2020)Song, Sohl-Dickstein, Kingma, Kumar, Ermon, and
  Poole]{song2020score}
Yang Song, Jascha Sohl-Dickstein, Diederik~P Kingma, Abhishek Kumar, Stefano
  Ermon, and Ben Poole.
\newblock Score-based generative modeling through stochastic differential
  equations.
\newblock \emph{arXiv preprint arXiv:2011.13456}, 2020.

\bibitem[Srivastava et~al.(2017)Srivastava, Valkov, Russell, Gutmann, and
  Sutton]{srivastava2017veegan}
Akash Srivastava, Lazar Valkov, Chris Russell, Michael~U Gutmann, and Charles
  Sutton.
\newblock Veegan: Reducing mode collapse in {GAN}s using implicit variational
  learning.
\newblock \emph{Advances in neural information processing systems}, 30, 2017.

\bibitem[Terrell \& Scott(1992)Terrell and Scott]{terrell1992variable}
George~R Terrell and David~W Scott.
\newblock Variable kernel density estimation.
\newblock \emph{The Annals of Statistics}, pp.\  1236--1265, 1992.

\bibitem[Tolman(1979)]{tolman1979principles}
Richard~Chace Tolman.
\newblock \emph{The principles of statistical mechanics}.
\newblock Courier Corporation, 1979.

\bibitem[Van~Kerm(2003)]{van2003adaptive}
Philippe Van~Kerm.
\newblock Adaptive kernel density estimation.
\newblock \emph{The Stata Journal}, 3\penalty0 (2):\penalty0 148--156, 2003.

\bibitem[Wand \& Jones(1994)Wand and Jones]{wand1994kernel}
Matt~P Wand and M~Chris Jones.
\newblock \emph{Kernel smoothing}.
\newblock CRC press, 1994.

\bibitem[Wang et~al.(2019)Wang, Tang, Bajaj, and Liu]{wang2019stein}
Dilin Wang, Ziyang Tang, Chandrajit Bajaj, and Qiang Liu.
\newblock Stein variational gradient descent with matrix-valued kernels.
\newblock \emph{Advances in neural information processing systems}, 32, 2019.

\bibitem[Wang \& Li(2022)Wang and Li]{wang2022accelerated}
Yifei Wang and Wuchen Li.
\newblock Accelerated information gradient flow.
\newblock \emph{Journal of Scientific Computing}, 90:\penalty0 1--47, 2022.

\bibitem[Wibisono(2018)]{wibisono2018sampling}
Andre Wibisono.
\newblock Sampling as optimization in the space of measures: The {L}angevin
  dynamics as a composite optimization problem.
\newblock In \emph{Conference on Learning Theory}, pp.\  2093--3027. PMLR,
  2018.

\end{thebibliography}
\bibliographystyle{iclr2024_conference}
\newpage
\appendix
\section{Derivation of updates for Gaussian}\label{app:updateGaussian}
In this section, we derive the closed form expressions for updating a Gaussian distribution, under the Ornstein-Uhlenbeck process. 

We begin with the derivation of \Cref{eq:1dGaussianrhoT}, which is the approximate Wasserstein proximal of the distribution at iteration $k$. In the following derivation, we discard constants that do not depend on $x$ and $y$ (but are allowed to depend on $\mu_k, \sigma_k^2$). We begin with computing the normalization constant of $K(x,y)$, given by the denominator of \Cref{eq:kernelDef}.
\begin{align*}
    & \int_{\R^d} \exp\left(-\frac{1}{2\beta}\left(V(z) + \frac{(z-y)^2}{2T}\right)\right) d z\\
    =& \int_{\R^d} \exp\left(-\frac{1}{2\beta}\left(\frac{az^2}{2} + \frac{(z-y)^2}{2T}\right)\right) d z\\
    =& \int_{\R^d} \exp\left(-\frac{1}{2\beta} \left(\frac{a}{2}+\frac{1}{2T}\right)\left(z - \frac{y/2T}{a/2 + 1/2T}\right)^2\right) \exp\left(\frac{1}{2\beta} \frac{(y/2T)^2}{a/2 + 1/2T}\right) \exp\left(\frac{-y^2}{4\beta T}\right) d z \\
    \propto& \exp \left(\frac{y^2}{2\beta} \left(\frac{1}{2T(1+aT)} - \frac{1}{2T}\right)\right).
\end{align*}

Substituting into the definition of $\rho_{k,T}$,
\begin{align*}
    &\rho_{k,T}(x) = \int_\R K(x,y) \frac{1}{\sqrt{2 \pi} \sigma_k} \exp\left(-\frac{(y-\mu_k)^2}{2\sigma_k^2}\right)\, d y  \\
    &\propto \int \exp\left(-\frac{1}{2\beta} \left(\frac{ax^2}{2} + \frac{(x-y)^2}{2T}\right)\right) \exp\left(-\frac{(y-\mu_k)^2)}{2\sigma_k^2} )-\frac{y^2}{2\beta} \left(\frac{1}{2T(1+aT)} - \frac{1}{2T}\right)\right)\, d y \\
    &= \exp\left(-\frac{ax^2}{4\beta}\right) \int\exp\left(-\frac{(y-x)^2}{4\beta T} - \frac{(y-\mu_k)^2}{2\sigma_k^2}-\frac{y^2}{2\beta} \left(\frac{1}{2T(1+aT)} - \frac{1}{2T}\right)\right) dy \\
    &= \exp\left(-\frac{ax^2}{4\beta}\right) \\
    & \qquad \times \int\exp \left( -\frac{1}{2} \left[y^2\left(\frac{1}{2\beta T(1+aT)} + \frac{1}{\sigma_k^2}\right) - 2y \left(\frac{x}{2\beta T} + \frac{\mu_k}{\sigma_k^2}\right) + \left(\frac{x^2}{2\beta T} + \frac{\mu_k^2}{\sigma_k^2}\right)\right] \right) d y \\
 &\propto \exp\left(-\frac{ax^2}{4\beta} - \frac{x^2}{4\beta T} + \frac{1}{2}\frac{(\frac{x}{2\beta T} + \frac{\mu_k}{\sigma_k^2})^2}{\frac{1}{2\beta T(1+aT)} + \frac{1}{\sigma_k^2}}\right) \\
 &\qquad \times \int \exp\left[-\frac{1}{2} \left(\frac{1}{2\beta T(1+aT)} + \frac{1}{\sigma_k^2}\right)\left(y - \frac{\frac{x}{2\beta T} + \frac{\mu_k}{\sigma_k^2}}{\frac{1}{2\beta T(1+aT)} + \frac{1}{\sigma_k^2}}\right)^2\right] d y.
\end{align*}

Observe in the final expression, the integral is of a Gaussian density whose variance does not depend on $x$, hence integrates to something independent of $x$. Hence, $\rho_{k,T} \sim \gN(\tilde\mu_{k+1}, \tilde\sigma_{k+1}^2)$ is a Gaussian density on $x$, with mean and variance
\begin{equation*}
    \tilde\mu_{k+1}
    = \frac{\mu_k}{1+aT},\quad \tilde\sigma_{k+1}^2 = \frac{\sigma_k^2}{(1+aT)^2} + \frac{2\beta T}{1+aT}.
\end{equation*}
This shows \Cref{eq:1dGaussianrhoT}. We now compute it in the multi-dimensional case as well, taking special care where the covariance matrices do not commute.

Computing the denominator of \Cref{eq:kernelDef} as before, we have
\begin{align*}
    & \int_{\R^d} \exp\left(-\frac{1}{2\beta}\left(V(z) + \frac{\|z-y\|^2}{2T}\right)\right) d z\\
    =& \int_{\R^d} \exp\left(-\frac{1}{2\beta}\left(\frac{z^\top \Sigma^{-1} z}{2} + \frac{\|z-y\|^2}{2T}\right)\right) d z\\
    =& \int_{\R^d} \Bigg[\exp\left(-\frac{1}{2\beta} \left(z - \left(\frac{\Sigma^{-1}}{2} + \frac{I}{2T}\right)^{-1}\frac{y}{2T}\right)^\top \left(\frac{\Sigma^{-1}}{2}+\frac{I}{2T}\right) \left(z - \left(\frac{\Sigma^{-1}}{2} + \frac{I}{2T}\right)^{-1}\frac{y}{2T}\right)\right) \\ 
    &\qquad \exp\left(\frac{1}{2\beta} \left(\frac{y}{2T}\right)^\top \left({\frac{\Sigma^{-1}}{2} + \frac{I}{2T}}\right)^{-1} \left(\frac{y}{2T}\right) \right) \exp\left(\frac{-\|y\|^2}{4\beta T}\right)\Bigg] d z \\
    \propto& \exp \left(\frac{1}{2\beta} y^\top\left(\left(2T(I+\Sigma^{-1}T)\right)^{-1} - \frac{I}{2T}\right) y\right),
\end{align*}
where the second equality follows from completing the square, and the final expression from integrating with respect to $z$, noting that the first exponential term is a Gaussian whose variance does not depend on $y$. We compute the approximate Wasserstein proximal $\rho_{k,T}$, given $\rho_{k,0} \sim \gN(\mu_k, \Sigma_k)$:
\begin{align*}
    \rho_{k,T}(x) &\propto \int_{\R^d} K(x,y) \exp\left(-\frac{1}{2}(y-\mu_k)^\top \Sigma_k^{-1} (y-\mu_k)\right)\, d y \\
    &=\int \exp \Bigg[ -\frac{1}{2\beta} \left(\frac{1}{2} x^\top \Sigma^{-1}x + \frac{\|y-x\|^2}{2T}\right) -\frac{1}{2} (y-\mu_k)^\top \Sigma_k^{-1} (y-\mu_k)\\
    & \hspace{1.7cm} - \frac{1}{2\beta} y^\top \left[\frac{1}{2T} (I + T\Sigma^{-1})^{-1} - \frac{I}{2T}\right]y\Bigg] dy \\
    &\propto \exp\left(-\frac{x^\top \Sigma^{-1} x}{4\beta} - \frac{\|x\|^2}{4\beta T}\right)\\
    & \quad \cdot \int \exp \Bigg[ y^\top\left(-\frac{I}{4\beta T} - \frac{\Sigma_k^{-1}}{2}- \frac{(I + T\Sigma^{-1})^{-1}}{4\beta T} + \frac{I}{4\beta T}\right) y  \\ 
    & \hspace{1.7cm} + y^\top \left(\frac{x}{4\beta T} + \frac{\Sigma_k^{-1}\mu_k}{2}\right) + \left(\frac{x}{4\beta T} + \frac{\Sigma_k^{-1}\mu_k}{2}\right)^\top y \Bigg] dy \\
    &\propto \exp \Bigg(-\frac{1}{2}\Bigg[x^\top \left(\frac{\Sigma^{-1}}{2\beta} + \frac{I}{2\beta T}\right) x  \\
    & \hspace{2.1cm}  - \left(\frac{x}{2\beta T} + \Sigma_k^{-1} \mu_k\right)^\top \left(\Sigma_k^{-1} + \frac{1}{2\beta T}(I + T\Sigma^{-1})^{-1}\right)^{-1} \left(\frac{x}{2\beta T} + \Sigma_k^{-1} \mu_k\right) \Bigg]\Bigg)\\ 
    &\propto \exp \Bigg(-\frac{1}{2} \Bigg[x^\top \left(\frac{\Sigma^{-1}}{2\beta} + \frac{I}{2\beta T} - \left[(2\beta T)^2\Sigma_k^{-1} + 2\beta T(I + T\Sigma^{-1})^{-1}\right]^{-1}\right) x \\
    & \hspace{2.1cm}  + x^\top\left[(2\beta T \Sigma_k^{-1} + (I + T\Sigma^{-1})^{-1})^{-1} \Sigma_k^{-1} \mu_k\right]\\
    & \hspace{2.1cm}  + \left[(2\beta T \Sigma_k^{-1} + (I + T\Sigma^{-1})^{-1})^{-1} \Sigma_k^{-1} \mu_k\right]^\top x\Bigg]\Bigg).
\end{align*}
The regularized Wasserstein proximal $\rho_{k,T}$ is thus Gaussian with mean and inverse covariance 
\begin{align*}
    \tilde{\Sigma}_{k+1}^{-1} &= \frac{\Sigma^{-1}}{2\beta} + \frac{I}{2\beta T} - \left[(2\beta T)^2\Sigma_k^{-1} + 2\beta T(I + T\Sigma^{-1})^{-1}\right]^{-1} \\
    &= \left[(2\beta T)^2\Sigma_k^{-1} + 2\beta T(I + T\Sigma^{-1})^{-1}\right]^{-1} \\
    &\qquad \left(\left[(2\beta T)^2\Sigma_k^{-1} + 2\beta T(I + T\Sigma^{-1})^{-1}\right]\left(\frac{\Sigma^{-1}}{2\beta} + \frac{I}{2\beta T}\right)- I\right)\\ 
    &= \left[(2\beta T)^2\Sigma_k^{-1} + 2\beta T(I + T\Sigma^{-1})^{-1}\right]^{-1}\\
    & \qquad \left(\left[(2\beta T)\Sigma_k^{-1} + (I + T\Sigma^{-1})^{-1}\right](T\Sigma^{-1} + I)- I\right)\\ 
    &= \left[(2\beta T)^2\Sigma_k^{-1} + 2\beta T(I + T\Sigma^{-1})^{-1}\right]^{-1} \\
    & \qquad (2\beta T \Sigma_k^{-1} (I + T\Sigma^{-1}))\\
    &= \left(2\beta T I + \Sigma_k (I + T\Sigma^{-1})^{-1} \right)^{-1}(I + T\Sigma^{-1});\\
    &= \left(2\beta T (I + T\Sigma^{-1})^{-1} + (I + T\Sigma^{-1})^{-1}\Sigma_k (I + T\Sigma^{-1})^{-1} \right)^{-1},
\end{align*}
\begin{align*}
    \tilde{\mu}_{k+1} &= \tilde\Sigma_{k+1}\left[(2\beta T \Sigma_k^{-1} + (I + T\Sigma^{-1})^{-1})^{-1} \Sigma_k^{-1} \mu_k\right] \\
    &= (I + T\Sigma^{-1})^{-1} \mu_k.
\end{align*}

This shows the recurrence relation \Cref{eqs:nDGaussianrhoT} for the distribution update under BRWP.

\section{Recurrence relation for eigenvalues for commuting Gaussians}\label{app:evalueGaussians}

We let $\Sigma = \diag(\xi^{(1)}, ..., \xi^{(d)})$ be positive definite, the stationary distribution $\Pi$ of the discrete scheme \Cref{eq:BackwardsWProxStep} be given by
    \begin{equation}
        \Pi \sim \gN(0, \Sigma_\infty),\, \Sigma_\infty = \diag( \tau_\infty^{(i)} \mid i=1,...,d),
    \end{equation}
    \begin{equation}
        \tau_\infty^{(i)} = \beta\xi^{(i)} (1-T^2 \xi^{(i)-2}), i=1,...,d.
    \end{equation}

Observe that the $i$-th entry of $\tilde\Sigma_{k+1}^{-1}$ is given by
\begin{align*}
    \left(\tilde\Sigma_{k+1}^{-1}\right)_{i,i} &= \left(2\beta T + \frac{\tau^{(i)}_k}{(1+T\xi^{(i) -1})}\right)^{-1} (1+T\xi^{(i) -1}), \cc
\end{align*}
and therefore 
\begin{align*}
    (I - \eta\Sigma^{-1} + \eta\beta\tilde\Sigma_{k+1}^{-1})_{i,i} &= 1 - \eta \xi^{(i)-1} + \eta\beta \left(2\beta T + \frac{\tau^{(i)}_k}{(1+T\xi^{(i)-1})}\right)^{-1} (1+T\xi^{(i)-1}) \cc\\ 
    & = 1 - \eta \xi^{(i)-1} + \frac{\eta\beta (1+T\xi^{(i)-1})^2}{\tau_k^{(i)}+2\beta T(1+T\xi^{(i)-1})}. \cc
\end{align*}

Temporarily dropping the $(i)$ superscripts that denote the coordinate, we consider the evolution of the covariance in the $i$-th coordinate, which is sufficient since the covariance matrices are diagonal. Indeed, from \Cref{eq:MultiDimGaussianVar} it evolves as
\begin{equation}\label{eq:MultivarGaussianCoordEvo}
    \tau_{k+1} = \left(1 - \eta \xi^{-1} + \frac{\eta\beta (1+T\xi^{-1})^2}{\tau_k+2\beta T(1+T\xi^{-1})}\right)^2\tau_k. \cc
\end{equation}

Observe that this is the same as \Cref{eq:WPKernelVarRecRel} up to a renaming of variables, in particular by letting $a = \xi^{-1}$ and $\sigma_k^2 = \tau_k$. We thus have the same fixed points, given by 
\begin{equation*}
    \tau_\infty = \beta \xi(1- T^2\xi^{-2}). \cc
\end{equation*}

We wish to consider the mixing time with respect to this variance. Consider the ansatz
\begin{equation*}
    \sqrt{\tau_{k}} = \sqrt{\beta \xi(1- T^2\xi^{-2})} + \sqrt{1+T\xi^{-1}}\gamma_k.
\end{equation*}

We compute a recurrence relation for $(\gamma_k)$ using \Cref{eq:MultivarGaussianCoordEvo}
\begin{align*}
    \sqrt{\tau_{k+1}} &= \sqrt{\tau_k}\left[1 + \eta\left(-\xi^{-1} + \frac{\beta(1+T\xi^{-1})^2}{(\sqrt{\beta\xi(1-T^2\xi^{-2})} + \sqrt{1+T\xi^{-1}}\gamma_k)^2+2\beta T(1+T\xi^{-1})}\right)\right] \cc\\
    &= \sqrt{\tau_k} \left[1+ \eta \left(-\xi^{-1} + \frac{\beta (1+T\xi^{-1})}{(\sqrt{\beta\xi(1-T\xi^{-1})} + \gamma_k)^2 + 2\beta T}\right)\right] \cc\\
    &= \sqrt{\tau_k} \left[1 + \eta \left(\frac{-\xi^{-1} ((\sqrt{\beta\xi(1-T\xi^{-1})} + \gamma_k)^2 + 2\beta T) + \beta(1+T\xi^{-1})}{(\sqrt{\beta\xi(1-T\xi^{-1})} + \gamma_k)^2 + 2\beta T}\right)\right] \cc\\
    &= \sqrt{\tau_k} \left[1 - \frac{\eta\gamma_k [2\xi^{-1}\sqrt{\beta\xi(1-T\xi^{-1})} + \xi^{-1}\gamma_k]}{(\sqrt{\beta\xi(1-T\xi^{-1})} + \gamma_k)^2 + 2\beta T}\right] \cc\\
    &= \sqrt{\tau_k} - \sqrt{\tau_k} \left[\frac{\eta\gamma_k [2\xi^{-1}\sqrt{\beta\xi(1-T\xi^{-1})} + \xi^{-1}\gamma_k]}{(\sqrt{\beta\xi(1-T\xi^{-1})} + \gamma_k)^2 + 2\beta T}\right]. \cc
\end{align*}
Subtracting $\sqrt{\beta \xi(1- T^2\xi^{-2})}$ from both sides and dividing by $\sqrt{1+T\xi^{-1}}$, we have
\begin{align*}
    \gamma_{k+1} &= \gamma_{k} - \frac{\sqrt{\tau_k}}{\sqrt{1+T\xi^{-1}}}\left[\frac{\eta\gamma_k [2\xi^{-1}\sqrt{\beta\xi(1-T\xi^{-1})} + \xi^{-1}\gamma_k]}{(\sqrt{\beta\xi(1-T\xi^{-1})} + \gamma_k)^2 + 2\beta T}\right] \cc\\
    &= \gamma_k - (\sqrt{\beta\xi(1-T\xi^{-1})}+\gamma_k)\left[ \frac{\eta\gamma_k [2\xi^{-1}\sqrt{\beta\xi(1-T\xi^{-1})} + \xi^{-1}\gamma_k]}{(\sqrt{\beta\xi(1-T\xi^{-1})} + \gamma_k)^2 + 2\beta T}\right] \cc\\
    &= \gamma_k \left[1- \eta\xi^{-1} \frac{[2\sqrt{\beta\xi(1-T\xi^{-1})} + \gamma_k][\sqrt{\beta\xi(1-T\xi^{-1})} + \gamma_k]}{(\sqrt{\beta\xi(1-T\xi^{-1})} + \gamma_k)^2 + 2\beta T}\right] \cc\\
    &= \gamma_k[1-\eta\xi^{-1} \omega_k].
\end{align*}
We now show that $\omega_k = \omega(\gamma_k)$, where $\omega:(-\sqrt{\beta\xi(1-T\xi^{-1})}, +\infty) \rightarrow (0, +\infty)$,
\begin{equation}
    \omega(\gamma) =  \frac{[2\sqrt{\beta\xi(1-T\xi^{-1})} + \gamma][\sqrt{\beta\xi(1-T\xi^{-1})} + \gamma]}{(\sqrt{\beta\xi(1-T\xi^{-1})} + \gamma)^2 + 2\beta T},\cc
\end{equation}
satisfies $\omega_k \in (\delta, \Delta]$ for some $\delta >0$ depending only on $\gamma_0$, and $\Delta$ depending only on $\xi$ and $T$. This will give us linear convergence of $\gamma_k$ to zero, as long as $\eta \xi^{-1}\le 1/\Delta$, and $T\xi^{-1}<1$. 

First observe that 
\begin{equation*}
    \sqrt{\beta\xi(1-T\xi^{-1})} + \gamma_k = \frac{\sqrt{\tau_k}}{\sqrt{1+T\xi^{-1}}} > 0.\cc
\end{equation*}

Considering the translation $\bar\omega(\gamma) = \omega(\gamma-\sqrt{\beta\xi(1-T\xi^{-1})}) : (0, +\infty)\rightarrow \R$, we can simplify
\begin{equation*}
    \bar\omega(\gamma) = \frac{\gamma(\gamma + \sqrt{\beta\xi(1-T\xi^{-1})})}{\gamma^2 + 2\beta T}.
\end{equation*}

$\bar\omega$ is maximized at 
\begin{equation*}
    \max_{\gamma>0} \bar\omega(\gamma) = \frac{1}{2}\left(\sqrt{\frac{\xi + T}{2T}} + 1\right) \eqqcolon \Delta(\xi, T),
\end{equation*} 
obtained at the critical point 
\begin{equation*}
    \gamma = \sqrt{\frac{4\beta^2 T^2}{\beta\xi(1-T\xi^{-1})} + 2\beta T} + \frac{2\beta T}{\sqrt{\beta\xi(1-T\xi^{-1})}}.
\end{equation*}


This shows that $\omega$ is bounded above by $\Delta$, and gives a closed form for $\Delta$ in terms of $\xi$ and $T$. To show that $\omega$ is bounded below, we note that as $\gamma \rightarrow -\sqrt{\beta\xi(1-T\xi^{-1})}$, $\omega \rightarrow 0^+$. Moreover, as $\gamma \rightarrow +\infty$, $\omega \rightarrow 1$. Therefore, under the assumption that $\gamma_k \rightarrow 0$, $\omega_k = \omega(\gamma_k)$ is bounded from below. Moreover, if the convergence is monotonic, then $\omega(\gamma_k)$ is bounded from below by $\delta = \min(\omega(\gamma_0), \omega(0)) > 0$.

As $\gamma_k \rightarrow 0$, we have that $\omega(\gamma_k) \rightarrow \omega(0)$, which takes the following form independent of $\beta$:
\begin{equation}
    \omega(0) = \frac{2(\xi - T)}{\xi + T}.
\end{equation}
Putting everything together, if $\eta \xi^{-1} \le 1/\Delta$, then $1-\eta\xi^{-1} \omega(\gamma_k) \in (0,1-\eta\xi^{-1}\delta)$, and thus we have that $\gamma_k \rightarrow 0$ linearly and monotonically. Moreover, the factor is $1-\eta\xi^{-1} \frac{2(\xi-T)}{\xi+T}$ (meaning that $\gamma_{k+1}/\gamma_k \rightarrow 1-\eta\xi^{-1} \frac{2(\xi-T)}{\xi+T}$ as $k \rightarrow \infty$)\cc. This is summarized in \Cref{thm:MixingTime}. The proof of the additional statements is as follows.

\begin{proof}[Proof of \Cref{thm:MixingTime}]
    The evolution of the covariance is given as above. For the linear convergence of $\tau_{k}^{(i)}$ to $\tau_\infty^{(i)}$, recall the ansatz
    \begin{equation*}
        \sqrt{\tau_{k}^{(i)}} = \sqrt{\tau_\infty^{(i)}} + \sqrt{1+T\xi^{(i)-1}} \gamma_k^{(i)}.
    \end{equation*}
    We have linear convergence of $\gamma_k^{(i)}$, given by
    \begin{equation*}
        \gamma_{k+1}^{(i)} = \gamma_k^{(i)}[1-\eta\xi^{(i)-1} \omega^{(i)}(\gamma_k^{(i)})],
    \end{equation*}
    where $\omega^{(i)}(\gamma_k) \in (\delta^{(i)}, \Delta^{(i)})$ for all $k$. Moreover, the linear convergence is with factor $[1-2\eta\xi^{(i)-1}(\xi-T)/(\xi+T)]$. Therefore, we have linear convergence of $\sqrt{\tau_k^{(i)}}$ to $\sqrt{\tau_\infty^{(i)}}$ with the same factor. Since $\sqrt{\tau_k^{(i)}} \rightarrow \sqrt{\tau_\infty^{(i)}}$ monotonically, we have that the sequence is bounded by $\max(\sqrt{\tau_0^{(i)}},\sqrt{\tau_\infty^{(i)}})$. Therefore, we also have linear monotonic convergence of $\tau_k^{(i)}$ to $\tau_\infty^{(i)}$, with the same factor:
    \begin{align*}
        \frac{\tau_{k+1}^{(i)}-\tau_\infty^{(i)}}{\tau_{k}^{(i)}-\tau_\infty^{(i)}} &= \frac{\sqrt{\tau_{k+1}^{(i)}}-\sqrt{\tau_\infty^{(i)}}}{\sqrt{\tau_{k}^{(i)}}-\sqrt{\tau_\infty^{(i)}}} \cdot \frac{\sqrt{\tau_{k+1}^{(i)}}+\sqrt{\tau_\infty^{(i)}}}{\sqrt{\tau_{k}^{(i)}}+\sqrt{\tau_\infty^{(i)}}} \rightarrow 1-2\eta\xi^{(i)-1} \frac{\xi-T}{\xi+T}.
    \end{align*}

    The bound on the total variation follows directly from \Cref{thm:GaussianTV}. The constants are 
    \begin{equation}
        C = \frac{3}{2} \max_{i=1,...,d} \left|\frac{\tau_0^{(i)}}{\tau_\infty^{(i)}}-1\right|;
    \end{equation}
    \begin{equation}
        c = \max_i \left\{1 - \eta\xi^{(i)-1} \min\left(\omega^{(i)}(\gamma_0^{(i)}), \frac{2(\xi^{(i)}-T)}{\xi^{(i)}+T}\right)\right\},
    \end{equation}
    \begin{equation}
        \omega^{(i)}(\gamma_0^{(i)}) = \frac{\tau_0^{(i)} + \sqrt{\tau_0^{(i)}} \beta\xi^{(i)} (1-T^2 \xi^{(i)-2})}{\tau_0^{(i)} + 2\beta T(1 + T\xi^{(i)-1})}.
    \end{equation}
\end{proof} 
\section{Proof of Lyapunov convergence for Gaussians}\label{app:convergenceGaussian}

Here, we demonstrate convergence of the Lyapunov function in \Cref{prop:frobeniusLyapunov}. We begin by differentiating with respect to time, using \Cref{eqs:ctsEvo}.

\begin{align*}
    & \frac{d}{dt} \Tr((\ttS_\infty^{-1} - \ttS_t^{-1})^2) \\
    & = 2\Tr(\frac{d}{dt}(\ttS_\infty^{-1} - \ttS_t^{-1})(\ttS_\infty^{-1} - \ttS_t^{-1})) \\
    &= 2\Tr(\ttS_t^{-1} \frac{d\ttS_{t}}{dt}\ttS_t^{-1} (\ttS_\infty^{-1} - \ttS_t^{-1})) \\
    &= -2\Tr(\ttS_t^{-1} K^{-1} \left[ (\ttS_\infty^{-1} - \ttS_t^{-1})\Sigma_t + \Sigma_t (\ttS_\infty^{-1} - \ttS_t^{-1})\right]K^{-1}\ttS_t^{-1} (\ttS_\infty^{-1} - \ttS_t^{-1}))\\
    &= -2\Tr( \left[ (\ttS_\infty^{-1} - \ttS_t^{-1})\Sigma_t + \Sigma_t (\ttS_\infty^{-1} - \ttS_t^{-1})\right]K^{-1}\ttS_t^{-1} (\ttS_\infty^{-1} - \ttS_t^{-1})\ttS_t^{-1} K^{-1}) \\
    &\overset{(*)}{=} -4 \Tr( \left[ (\ttS_\infty^{-1} - \ttS_t^{-1})\Sigma_t\right]K^{-1}\ttS_t^{-1} (\ttS_\infty^{-1} - \ttS_t^{-1})\ttS_t^{-1} K^{-1}) \\
    &= -4 \Tr(  (\ttS_\infty^{-1} - \ttS_t^{-1})(K \ttS_t K - 2\beta T K) K^{-1}\ttS_t^{-1} (\ttS_\infty^{-1} - \ttS_t^{-1})(\ttS_t^{-1} K^{-1}) \\
    &= -4 \Tr(  (\ttS_\infty^{-1} - \ttS_t^{-1})(K - 2\beta T \ttS_t^{-1}) (\ttS_\infty^{-1} - \ttS_t^{-1})(\ttS_t^{-1} K^{-1})). \cc
\end{align*}

In $(*)$, we used that $\Tr(AB) = \Tr(A^\top B^\top)$. We now aim to bound the final term in the product, $\ttS_t^{-1} K^{-1}$. To do this, we use the following results from linear algebra.


\begin{proposition}\label{prop:specCloseToI}
    Suppose $A$ is Hermitian and positive definite, and $B$ is square of the same dimensions, satisfying:
    \begin{equation}
        x^* B x > \frac{1}{2}x^* A x, \quad\forall x \ne 0.
    \end{equation}
    Then $\rho(I - B^{-1}A) < 1$.
\end{proposition}
\begin{proof}
    Note that the positivity condition gives that $B$ is invertible. The eigenvalues of $I - B^{-1}A$ and $I - A^{1/2} B^{-1} A^{1/2}$ are equal. From the quadratic form inequality, we have for any (complex) $z \ne 0$,
    \[\Re(z^* A^{-1/2} B A^{-1/2} z) > \frac{1}{2} z^* z.\]
    Therefore the real part of each eigenvalue of $A^{-1/2} B A^{-1/2}$ satisfies $\Re \lambda_i(A^{-1/2} B A^{-1/2}) > 1/2$.

    Now note that $1-1/z$ is a conformal mapping, taking the half-plane $\Re(z) > 1/2$ to the unit disk $|w| < 1$. Thus the spectrum of $I - A^{1/2} B^{-1} A^{1/2}$ lies in the unit disk and we conclude.
\end{proof}

 Using \Cref{prop:specCloseToI} with $A = 4\beta T K^{-1}$ and $B = \ttS_t$, we satisfy the assumptions of the proposition, since:

\begin{equation}
    \ttS_t = K^{-1}\Sigma_t K^{-1} + 2\beta T K^{-1} \succeq \frac{1}{2} 4\beta T K^{-1}.
\end{equation}

Thus, we have the following bound on the spectral radius,

\begin{align*}
    &\rho(I - 4\beta T \ttS_t^{-1} K^{-1}) < 1 \\
    \Rightarrow &\rho(\frac{I}{4\beta T} - \ttS_t^{-1} K^{-1}) < \frac{1}{4\beta T}.
\end{align*}

In fact, we can conclude slightly more if we can bound the maximum and minimum eigenvalues of $\ttS_t$. Since we can bound $I\preceq K \preceq (1+c)I$, where $T = c\lambda_{\min}(\Sigma)$ with $c \in (0,1)$, we have that the minimum and maximum eigenvalues of $K^{-1/2} \ttS_t^{-1} K^{-1/2}$ are bounded as

\begin{equation*}
    \frac{1}{1+c}\lambda_{\min}(\ttS_t) \le \lambda_{\min}(K^{-1/2} \ttS_t K^{-1/2}) \le \lambda_{\max}(K^{-1/2} \ttS_tK^{-1/2}) \le \lambda_{\max}(\ttS_t).
\end{equation*}

Therefore, a bound for the eigenvalues of $I - \ttS_t^{-1} K^{-1}$ is given by

\begin{gather}
    \lambda_{\max}(\frac{I}{4\beta T} - \ttS_t^{-1} K^{-1}) = \frac{1}{4\beta T}-\frac{1}{\lambda_{\max}(K^{-1/2} \ttS_t K^{-1/2})} \le \frac{1}{4\beta T}-\frac{1}{\lambda_{\max}(\ttS_t)}, \\
    \lambda_{\min}(\frac{I}{4\beta T} - \ttS_t^{-1} K^{-1}) = 
    \frac{1}{4\beta T}-\frac{1}{\lambda_{\min}(K^{-1/2} \ttS_t K^{-1/2})} \ge \frac{1}{4\beta T}-\frac{1+c}{\lambda_{\min}(\ttS_t)},
\end{gather}

\begin{equation}
    \rho(\frac{I}{4\beta T} - \ttS_t^{-1} K^{-1}) = \max(|\lambda_{\min}|, |\lambda_{\max}|) < \frac{1}{4\beta T}. 
\end{equation}

To turn this bound on $\ttS_t^{-1}K^{-1}$ to a bound on the derivative of the Lyapunov function, we need the following trace inequality \citep{horn2012matrix,baumgartner2011inequality}.
\begin{proposition}[H\"older's inequality for trace]
    Let $A, B$ be (complex) square matrices , with absolute values $|A| = (A^* A)^{1/2},|B| = (B^* B)^{1/2}$. Then for $1\le p,q\le \infty$ satisfying $p^{-1} + q^{-1} = 1$, the trace inequality holds:
    \begin{equation}
        |\Tr(A^*B)| \le \|A\|_p \|B\|_q,
    \end{equation}
    where $\|\cdot\|_p$ are the Schatten $p$-norms defined as follows, where $\sigma_i$ are the singular values
    \begin{equation*}
    \begin{cases}
        \|A\|_p^p = \sum \sigma_i^p(A),&\, 1\le p<\infty,\\
        \|A\|_\infty = \sup \sigma_i(A),  &\, p=\infty.
    \end{cases}
    \end{equation*}

    In particular, if $A$ is symmetric and positive-definite, we have, where $\rho$ denotes the spectral radius,
    \[|\Tr(AB)| \le \|A\|_1 \|B\|_\infty = \Tr(A) \rho(B).\]
\end{proposition}

Using these two results, we expand the derivative of the Lyapunov function, noting that $(\ttS_\infty^{-1} - \ttS_t^{-1})(K - 2\beta T \ttS_t^{-1}) (\ttS_\infty^{-1} - \ttS_t^{-1})$ is positive definite (since $K - 2\beta T \ttS_t^{-1}\succeq 0$):

\begin{align*}
    &\frac{d}{dt} \Tr((\ttS_\infty^{-1} - \ttS_t^{-1})^2) \\
    =& -4 \Tr(  (\ttS_\infty^{-1} - \ttS_t^{-1})(K - 2\beta T \ttS_t^{-1}) (\ttS_\infty^{-1} - \ttS_t^{-1})(\ttS_t^{-1} K^{-1})) \\
    = &-4 \Tr(  (\ttS_\infty^{-1} - \ttS_t^{-1})(K - 2\beta T \ttS_t^{-1}) (\ttS_\infty^{-1} - \ttS_t^{-1})(\ttS_t^{-1} K^{-1} - \frac{I}{4\beta T})) \\
    &  - 4\Tr(  (\ttS_\infty^{-1} - \ttS_t^{-1})(K - 2\beta T \ttS_t^{-1}) (\ttS_\infty^{-1} - \ttS_t^{-1})(\frac{I}{4\beta T})) \\
    \le & \quad 4\Tr(  (\ttS_\infty^{-1} - \ttS_t^{-1})(K - 2\beta T \ttS_t^{-1}) (\ttS_\infty^{-1} - \ttS_t^{-1})) \rho(\ttS_t^{-1} K^{-1} - \frac{I}{4\beta T}) \\
    & - 4\Tr(  (\ttS_\infty^{-1} - \ttS_t^{-1})(K - 2\beta T \ttS_t^{-1}) (\ttS_\infty^{-1} - \ttS_t^{-1}))(\frac{1}{4\beta T}) \\
     \le &- 4\Tr(  (\ttS_\infty^{-1} - \ttS_t^{-1})(K - 2\beta T \ttS_t^{-1}) (\ttS_\infty^{-1} - \ttS_t^{-1})) (\frac{1}{4\beta T} - \rho(\frac{I}{4\beta T} - \ttS_t^{-1} K^{-1})) \\
    <&\ 0.
\end{align*}
This bound can be slightly refined to yield linear convergence. If $(\frac{1}{4\beta T} - \rho(\frac{I}{4\beta T} - \ttS_t^{-1} K^{-1}))$ is bounded away from zero and $K - 2\beta T\ttS_t^{-1}$ has (positive) eigenvalues also bounded away from zero, then we have linear convergence of $\|\ttS_\infty^{-1} - \ttS_t^{-1}\|_F^2$ to zero. Both of these assumptions can be justified using the equivalence of matrix norms, and bootstrapping convergence of $\ttS_t$ to $\ttS_\infty$.

\end{document}